\documentclass[letterpaper, 10 pt, conference]{ieeeconf}  % Comment this line out if you need a4paper

\IEEEoverridecommandlockouts                              % This command is only needed if 
\usepackage{times}
\usepackage[maxbibnames=6,maxcitenames=1,backend=biber,natbib=true, sorting=none, style=numeric-comp, giveninits,doi=false,isbn=false,url=false,eprint=false]{biblatex}

\usepackage{multicol}
\usepackage[acronym]{glossaries}
\usepackage{amsmath}
\usepackage{amssymb}
\usepackage{bm}
\usepackage{mathtools}
\usepackage{multirow}
\usepackage{subcaption}
\usepackage{comment}
\usepackage{setspace}
\usepackage{graphicx}

\usepackage{savesym}
\savesymbol{checkmark}
\savesymbol{labelindent}

\usepackage{dingbat}
\usepackage[table, dvipsnames]{xcolor}
\usepackage[inline]{enumitem}
\usepackage{array}
\usepackage{wrapfig}
\usepackage[hidelinks]{hyperref}
\usepackage{colortbl}
\usepackage{booktabs}
\usepackage{balance}
\usepackage{stfloats}
\usepackage[utf8]{inputenc}

\usepackage{caption}
\usepackage{tikz}
\usetikzlibrary{automata,arrows,positioning,matrix,shapes.geometric,shapes.misc, arrows.meta}

\glsdisablehyper

\newacronym{acnmp}{ACNMP}{Adaptive Conditional Neural Movement Primitive}
\newacronym{ba}{BA}{Bayesian Aggregation}
\newacronym{bc}{BC}{boundary condition}
\newacronym{clv}{CLV}{conditional latent variable}
\newacronym{cnmp}{CNMP}{Conditional Neural Movement Primitive}
\newacronym{cnn}{CNNs}{convolutional neural networks}
\newacronym{cnp}{CNP}{Conditional Neural Processes}
\newacronym{dmp}{DMP}{Dynamic Movement Primitive}
\newacronym{dof}{DoF}{Degree of Freedom}
\newacronym{gp}{GPs}{Gaussian Processes}
\newacronym{idmp}{IDMP}{Integral form of Dynamic Movement Primitive}
\newacronym{ik}{IK}{inverse kinematics}
\newacronym{il}{IL}{Imitation Learning}
\newacronym{ma}{MA}{Mean Aggregation}
\newacronym{mp}{MP}{Movement Primitive}
\newacronym{mse}{MSE}{mean squared error}
\newacronym{ndp}{NDP}{Neural Dynamic Policies}
\newacronym{nmp}{ProDMP}{Probabilistic Dynamic Movement Primitive}
\newacronym{nn}{NN}{neural networks}
\newacronym{np}{NP}{Neural Processes}
\newacronym{ode}{ODE}{ordinary differential equation}
\newacronym{pdmp}{ProDMP}{Probabilistic Dynamic Movement Primitive}
\newacronym{ppo}{PPO}{Proximal Policy Optimization}
\newacronym{promp}{ProMP}{Probabilistic Movement Primitive}
\newacronym{rl}{RL}{Reinforcement Learning}
\newacronym{sse}{SSE}{Summed Squared Error}

\newcommand{\obslat}{\bm{r}_m}
\newcommand{\obsunc}{\bm{\sigma}^2_{\bm{r}_m}}
\newcommand{\zpriormu}{\bm{\mu}_{\bm{z}, 0}}
\newcommand{\zpriorvar}{\bm{\sigma}_{\bm{z},0}^2}

\newcommand{\ipb}{\bm{\Phi}}
\newcommand{\ivb}{\dot{\bm{\Phi}}}
\newcommand{\bipb}{\bm{\Psi}}
\newcommand{\bivb}{\dot{\bm{\Psi}}}

\newcommand{\ipbt}{\bm{\Phi}^\intercal}
\newcommand{\ivbt}{\dot{\bm{\Phi}}^\intercal}
\newcommand{\ipbb}{\bm{\Phi}_b}
\newcommand{\ivbb}{\dot{\bm{\Phi}}_b}
\newcommand{\bipbb}{\bm{\Psi}_b}
\newcommand{\bivbb}{\dot{\bm{\Psi}}_b}

\newcommand{\ipbbt}{\bm{\Phi}^\intercal_b}
\newcommand{\ivbbt}{\dot{\bm{\Phi}}^\intercal_b}
\title{\LARGE \bf
\acrshortpl{pdmp}: A Unified Perspective on Dynamic and Probabilistic Movement Primitives}

\author{Ge Li$^{1}$, Zeqi Jin$^{1}$, Michael Volpp$^{1}$, Fabian Otto$^{2,3}$, Rudolf Lioutikov$^{1}$, and Gerhard Neumann$^{1}$% <-this % stops a space
\thanks{$^{1}${Karlsruhe Institute of Technology, Germany. ge.li@kit.edu}
        {\tt\small}}%
\thanks{$^{2}$Bosch Center for Artificial Intelligence, Germany.
        {\tt\small}}
\thanks{$^{3}$University of Tübingen, Germany.
        {\tt\small}}%
}

\begin{document}

\maketitle
\thispagestyle{empty}
\pagestyle{empty}

%%%%%%%%%%%%%%%%%%%%%%%%%%%%%%%%%%%%%%%%%%%%%%%%%%%%%%%%%%%%%%%%%%%%%%%%%%%%%%%%
\begin{abstract}
\acrfullpl{mp} are a well-known concept to represent and generate modular trajectories. 
\acrshortpl{mp} can be broadly categorized into two types: 
\begin{enumerate*}[label=(\alph*)]
\item dynamics-based approaches that generate smooth trajectories from any initial state, e.\,g., \acrfullpl{dmp}, and
\item probabilistic approaches that capture higher-order statistics of the motion, e.\,g., \acrfullpl{promp}.
\end{enumerate*}  
To date, however, there is no method that unifies both, i.\,e. that can generate smooth trajectories from an arbitrary initial state while capturing higher-order statistics.
In this paper, we introduce a unified perspective of both approaches by solving the ODE underlying the \acrshortpl{dmp}. 
We convert expensive online numerical integration of \acrshortpl{dmp} into basis functions that can be computed offline. 
These basis functions can be used to represent trajectories or trajectory distributions similar to \acrshortpl{promp} while maintaining all the properties of dynamical systems. 
Since we inherit the properties of both methodologies, we call our proposed model \acrfullpl{pdmp}. 
Additionally, we embed \acrshortpl{pdmp} in deep neural network architecture and propose a new cost function for efficient end-to-end learning of higher-order trajectory statistics. 
To this end, we leverage \acrlong{ba} for non-linear iterative conditioning on sensory inputs.
Our proposed model achieves smooth trajectory generation, goal-attractor convergence, correlation analysis, non-linear conditioning, and online re-planing in one framework. 
\end{abstract}
% \keywords{Movement Primitives, Probabilistic Learning, Imitation Learning}  

\section{Introduction}
\label{sec:introduction}

\acrfullpl{mp} are a prominent tool for motion representation and synthesis in robotics. 
They serve as basic movement elements, modulate the motion behavior, and form more complex movements through combination or concatenation. This work focuses on trajectory-based movement representations \cite{schaal2006dynamic, paraschos2013probabilistic}. Given a parameter vector, such representations generate desired trajectories for the robot to follow.
These methods have gained much popularity in imitation and reinforcement learning (IL, RL) \cite{maeda2017probabilistic, zhou2019learning, celik2022specializing, maeda2017phase, ottodeep} due to their concise parametrization and flexibility to modulate movement.
Current methods can be roughly classified into approaches based on dynamical systems \cite{schaal2006dynamic, ijspeert2013dynamical, gams2018deep, ridge2020training, bahl2020neural} and probabilistic approaches \cite{paraschos2013probabilistic, paraschos2018using, seker2019conditional}, with both types offering their own advantages.
The dynamical systems-based approaches, such as \acrfullpl{dmp}, guarantee that the generated trajectories start precisely at the current position and velocity of the robot, which allows for smooth trajectory replanning i.\,e., changing the parameters of the \acrshortpl{mp} during motion execution \cite{bahl2020neural, brandherm2019learning}. 
However, since \acrshortpl{dmp} represent the trajectory via the forcing term instead of a direct representation of the trajectory position, numerical integration from acceleration to position has to be applied to formulate the trajectory, which constitutes an additional workload and makes the estimation of the trajectory statistics difficult \cite{saveriano2021dynamic}.
Probabilistic methods, such as \acrfull{promp}, are able to acquire such statistics, thus making them the key enablers for acquiring variable-stiffness controllers and the trajectory's temporal and \acrshortpl{dof} correlation.
These methods further perform as generative models, facilitating the sampling of new trajectories. However, the lack of internal dynamics of these approaches suffers from discontinuities in position and velocity between old and new trajectories in the case of replanning.

In this work, we propose \acrfullpl{pdmp} which unify both methodologies. We show that the trajectory of a \acrshort{dmp}, obtained by integrating its second-order dynamical system, can be expressed by a linear basis function model that depends on the parameters of the \acrshort{dmp}, i.\,e., the weights of the forcing function and the goal attractor. 
The linear basis functions can be obtained by integrating the original basis functions used in the \acrshort{dmp} - an operation that only needs to be performed once offline in the \acrshortpl{pdmp}. 
Recently, \acrshort{mp} research has been extended to deep neural network (NN) architectures \cite{ridge2020training, bahl2020neural, seker2019conditional} that enable conditioning the trajectory generation on high-dimensional context variables, such as images.
Following these ideas, we integrate our representation into a deep neural architecture that allows non-linear conditioning on a varying number of conditioning events. 
These events are aggregated using \acrfull{ba} into a latent probabilistic representation \cite{volpp2020bayesian} which is mapped to a Gaussian distribution in the parameter space of the \acrshortpl{pdmp}.
We summarize the contributions of this paper as: 
\begin{enumerate*}[label=(\alph*)]
    \item We unify \acrshortpl{promp} and \acrshortpl{dmp} into one consistent framework that inherits the benefits of both formulations.
    \item We enable to compute distributions and to capture correlations of \acrshortpl{dmp} trajectories, while
    \item the robot's current state can be inscribed into the trajectory distribution through boundary conditions, allowing for smooth replanning.
    \item Moreover, the offline integration of the basis functions significantly facilitates the integration into neural network architectures, reducing the computation time by a factor of 10. 
    \item Hence, we embed \acrshortpl{pdmp} in a deep encoder-decoder architecture that allows non-linear conditioning on a set of high-dimensional observations with varying information levels.
\end{enumerate*}
We evaluate our method on three digit-writing tasks using images as inputs, a simulated robot pushing task with a complex physical interaction, and a real robot picking task with shifting object positions. 
We compare our model with state-of-the-art NN-based \acrshortpl{dmp} \cite{gams2018deep, ridge2020training, bahl2020neural} and the NN-based probabilistic method \cite{seker2019conditional}. 

%%%%%%%%%%%%%%%%%%%%%%%%%%%%%%%%%%%%%%%%%%%%%%%%%%%%%%
%
%    Related work
%
%%%%%%%%%%%%%%%%%%%%%%%%%%%%%%%%%%%%%%%%%%%%%%%%%%%%%%

\section{Related Work}
\citet{paraschos2013probabilistic} established \acrshortpl{promp} to model \acrshortpl{mp} as a trajectory distribution that captures temporal correlation and correlations between the \acrshortpl{dof}.
\acrshortpl{promp} maintain a Gaussian distribution over the parameters and can map it to the corresponding trajectory distribution using a linear basis function model. In contrast, such a distribution mapping is not allowed for the \acrshort{dmp}-based approaches, as the trajectory is integrated numerically from the forcing function.
% due to the numerical integration in the dynamical system. 
Previous methods, like GMM/GMR-DMPs~\cite{calinon2012statistical,yang2018robot} used Gaussian Mixture Models to cover the trajectories' domain. 
Yet, this does not capture temporal correlation nor does it provide a generative approach to trajectories. 
% It is also unclear how to integrate such a GMR approach in a deep NN architecture which has recently been used in both dynamic and probabilistic \acrshortpl{mp} representations.
Other methods which have learned distributions over \acrshort{dmp} weights \cite{meier2016probabilistic, amor2014interaction}, do not connect the weights distribution to the trajectory distribution, as trajectories can only be obtained by integration. Hence, it is also hard to learn the weights distribution reversely from trajectories in an end-to-end manner.

To learn \acrshort{dmp} parameters from high-dimensional sensory inputs, \citet{gams2018deep, ridge2020training} designed an encoder-decoder architecture to learn the weights of a single \acrshort{dmp} from digit images, and derive the gradient of the trajectory with respect to the learnable parameters. \citet{bahl2020neural} propose \acrfull{ndp} that allow replanning the \acrshort{dmp} parameters throughout the execution of the trajectory which has also been extended to the RL setting.
The learning objective of these two methods for IL is to optimize the \acrfull{mse} between the predicted and the ground-truth trajectories using backpropagation. 
However, to formulate a trajectory, \acrshortpl{dmp} must apply numerical integration during the NN training procedure, which significantly increases the computational workload in both forward and backward propagation, rendering these approaches cumbersome to use.
% which will lead to a speed-precision dilemma. 
% A small $\mathrm{d}t$ leads to high precision and control frequency but increases iteration steps and workload in both forward and backward propagation. 
Additionally, the integration-based trajectory representation limits the use of probabilistic methods and hence these NN-\acrshortpl{dmp} approaches cannot be trained using a probabilistic log-likelihood (LL) loss. 

Probabilistic \acrshortpl{mp} approaches have also been extended with deep \acrshort{nn} architectures.
% to support non-linear conditioning.
\citet{seker2019conditional} directly use a \acrlong{cnp} model \cite{garnelo2018conditional} as a trajectory generator, i.e. \acrfullpl{cnmp}, to predict the trajectory distribution with an encoder-decoder \acrshortpl{nn} model.
While such an architecture enables non-linear conditioning on high-dimensional inputs, it can only predict an isotropic trajectory variance at each time step. The temporal and \acrshortpl{dof} correlations are missing, which makes sampling consistently in time and \acrshortpl{dof} infeasible.
Besides, both \acrshortpl{promp} and \acrshortpl{cnmp} neglect dynamics, i.\,e.\,, when changing trajectory parameters during trajectory execution, the newly generated trajectory will contain discontinuities at the replanning time point. 
To execute such trajectories, a heuristic controller is used to freeze the time and catch up with the jump \cite{seker2019conditional}. However, such a waiting mechanism does not scale to time-sensitive motions and tasks.

\section{A Unified Perspective on Dynamic and Probabilistic Movement Primitives}
\label{sec:idmp}
We first briefly cover the fundamental aspects of \acrshortpl{dmp}. Then, we derive the analytical solution of the \acrshortpl{dmp}’ \acrshort{ode} to develop our new \acrshortpl{pdmp} representation. For convenience, we introduce our approach through a 1-\acrshort{dof} dynamical system and later extend it to a high-\acrshortpl{dof} system.
%%%%%%%%%%%%%%%%%%%%%%%%%%%%%%%%%%%%%%%%%%%%%%%%%%%%%%%%%%%%%%%%%%%%%%%%%%
% DMP
%%%%%%%%%%%%%%%%%%%%%%%%%%%%%%%%%%%%%%%%%%%%%%%%%%%%%%%%%%%%%%%%%%%%%%%%%%
\subsection{Solving \acrshortpl{dmp}' Ordinary Differential Equation}
\label{sbsec:dmp_ode}
For a single movement execution as a trajectory $\bm{\lambda} = [y_t]_{t=0:T}$, \citet{schaal2006dynamic, ijspeert2013dynamical} model it as a second-order linear dynamical system with a non-linear forcing function $f$,
\par\nobreak\vspace{-0.3cm}
{\small
\begin{equation}
    \tau^2\ddot{y} = \alpha(\beta(g-y)-\tau\dot{y})+ f(x), \quad f(x) = x\frac{\sum\varphi_i(x)w_i}{\sum\varphi_i(x)} = x\bm{\varphi}_x^\intercal\bm{w},
    \label{eq:dmp_original}
\end{equation}
}%
where $y = y(t),~\dot{y}=\mathrm{d}y/\mathrm{d}t,~\ddot{y} =\mathrm{d}^2y/\mathrm{d}t^2$ represent the position, velocity, and acceleration of the system at time step $t$, respectively.
Here, we use the original formulation of \acrshortpl{dmp} in \cite{schaal2006dynamic} without any extensions. $\alpha$ and $\beta$ are spring-damper constants, 
$g$ is a goal attractor, and $\tau$ is a time constant which can be used to adapt the execution speed of the resulting trajectory. To this end, \acrshortpl{dmp} define the forcing function over an exponentially decaying phase variable $x(t)=\exp(-\alpha_x/\tau \; t)$, where  $\varphi_i(x)$ represents the (unnormalized) basis functions and $w_i \in \bm{w}$, $i=1...N$ are the corresponding weights. 
% %Due to the phase term $x$, the forcing function will asymptotically go to $0$ and, hence, the system dynamics is dominated by the stable goal attractor for $t \rightarrow \infty$. 
The trajectory of the motion $\bm{\lambda}$ is obtained by integrating the dynamical system, or more specifically, numerical integration from starting time to the target time point. 
% %In contrast to probabilistic approaches, \reb{such as \acrshortpl{promp}}, \acrshortpl{dmp} inherently allow for replanning \cite{brandherm2019learning}, i.\,e., changing the weights and goal attractor during trajectory execution without introducing jumps into positions or velocities. 
The dynamical system defined in Eq.~(\ref{eq:dmp_original}) is a second-order linear non-homogeneous \acrfull{ode} with constant coefficients, whose closed-form solution can be derived analytically. 
We rewrite the \acrshort{ode} and its homogeneous counterpart in standard form as
\par\nobreak\vspace{-0.3cm}
{\small
\begin{align}
    \textbf{\acrshortpl{dmp}' ODE:}~~\ddot{y} + \frac{\alpha}{\tau}\dot{y} + \frac{\alpha\beta}{\tau^2} y &=\frac{f(x)}{\tau^2}+\frac{\alpha \beta}{\tau^2} g \equiv F(x, g), \label{eq:dmp_non_homo}\\
    \textbf{Homo. ODE:}~~\ddot{y} + \frac{\alpha}{\tau}\dot{y} + \frac{\alpha\beta}{\tau^2} y &= 0,
    \label{eq:dmp_homo}
\end{align}
}%
where $F$ denotes some function of $x$ and $g$. Using the method of variation of constants \cite{teschl2012ordinary}, the closed-form solution of the second-order \acrshort{ode} in Eq.~(\ref{eq:dmp_non_homo}), i.\,e.\,, the trajectory position, is
\par\nobreak\vspace{-0.1cm}
{\small
\begin{equation}
    y = c_1y_1 + c_2y_2 -y_1 \int\frac{y_2F}{Y}\mathrm{d}t + y_2\int\frac{y_1F}{Y}\mathrm{d}t,
    \label{eq:dmp_closed_form_position}
\end{equation}
}%
where $y_1$, $y_2$ are two linearly independent complementary functions of the homogeneous ODE given in Eq.~(\ref{eq:dmp_homo}). $c_1$, $c_2$ are two constants which are determined by the \acrfullpl{bc} of the ODE, and $Y = y_1 \dot{y}_2 - \dot{y}_1y_2$. Both integrals in Eq.~(\ref{eq:dmp_closed_form_position}) are indefinite. 
With appropriate values $\beta = \alpha/4$ \cite{schaal2006dynamic, ijspeert2013dynamical}, the system is critically damped, and the corresponding characteristic equation of the homogeneous ODE, i.\,e.\, $\Delta = (\alpha^2-4\alpha\beta)/\tau^2$ will be $0$. 
Consequently, $y_1$, $y_2$ will take the form
\par\nobreak\vspace{-0.4cm}
{\small
\begin{equation}
    y_1 = y_1(t) = \exp\left(-\frac{\alpha}{2\tau}t\right),~
    y_2 = y_2(t) = t\exp\left(-\frac{\alpha}{2\tau}t\right).
    \label{eq:dmp_y1_y2}
\end{equation}
}%
Using this result, the term $Y$ can also be simplified to $Y = \exp{(-\alpha t/\tau)}\neq0.$ 
To get $y$, we need to solve the two indefinite integrals in Eq.~(\ref{eq:dmp_closed_form_position}) as\\
\par\nobreak\vspace{-0.7cm}
{\small
\begin{align}
    \begin{aligned}
        I_1(t) &= \int\frac{y_2F}{Y}\mathrm{d}t = \int t\exp\left(\frac{\alpha}{2\tau}t\right) F(x, g)\mathrm{d}t,\\
        I_2(t) &= \int\frac{y_1F}{Y}\mathrm{d}t =\int\exp\left(\frac{\alpha}{2\tau}t\right)F(x,g)\mathrm{d}t.    \label{eq:dmp_i1_i2}
    \end{aligned}
\end{align}
}%
\begin{spacing}{1.1}
\noindent Applying the Fundamental Theorem of Calculus, i.\,e.\,, $\int h(t)\mathrm{d}t = \int_0^th(t')\mathrm{d}t' + c$, $c \in \mathbb R$, together with the definition of the forcing function $f$ in Eq.~(\ref{eq:dmp_original}) and $F(x,g)$ in Eq.~(\ref{eq:dmp_non_homo}), $I_1(t)$ can be expressed as
\end{spacing}
\par\nobreak\vspace{-0.3cm}
{\small
\begin{align}
    &\begin{aligned}
        I_1(t) =\frac{1}{\tau^2}\bigg[\int_0^t t'\exp\Big(\frac{\alpha}{2\tau}&t'\Big)x(t') \bm{\varphi}_x^\intercal\bm{w}\mathrm{d}t'\\
        +& \int_0^t t'\exp\Big(\frac{\alpha}{2\tau}t'\Big)\frac{\alpha^2}{4} g \mathrm{d}t'\bigg] + c_3 \label{eq:dmp_i1_int}
    \end{aligned}\\
    &\begin{aligned}
      \phantom{I_1(t)} =\frac{1}{\tau^2}\bigg[\int_0^t t'\exp&\Big(\frac{\alpha}{2\tau}t'\Big)x(t')\bm{\varphi}_x^\intercal \mathrm{d}t'\bigg]\bm{w}\\
       +& \bigg[\Big(\frac{\alpha}{2\tau}t-1\Big)\exp\Big(\frac{\alpha}{2\tau}t\Big) + 1\bigg]g + c_3 \label{eq:dmp_i1_decouple},
    \end{aligned}
\end{align}
}%
\noindent where $c_3$ is a constant fixed by the \acrshortpl{bc}. From Eq.~(\ref{eq:dmp_i1_int}) to Eq.~(\ref{eq:dmp_i1_decouple}), we move the time-independent parameters $\bm{w}$ and $g$ out of their corresponding integrals.
Notice that the remaining part of the second integral has an analytical solution. The remaining part of the first integral, however, has no closed-form solution because the basis functions $\bm{\varphi}_x$ may be arbitrarily complex.
Denoting these integrals as
\par\nobreak\vspace{-0.4cm}
{\small
\begin{align}
    \begin{aligned}
        \bm{p}_1(t) &\equiv \frac{1}{\tau^2}\int_0^t t'\exp\Big(\frac{\alpha}{2\tau}t'\Big)x(t')\bm{\varphi}_x^\intercal \mathrm{d}t',\\
    q_1(t) &\equiv \Big(\frac{\alpha}{2\tau}t  - 1\Big)\exp\Big(\frac{\alpha}{2\tau}t\Big) +1 \label{eq:dmp_p1_q1},    
    \end{aligned}
\end{align}
}%
where $\bm{p_1}$ is a N-dim vector and $q_1$ a scalar, we can express
$I_1(t) = \bm{p}_1(t)^\intercal\bm{w} + q_1(t)g + c_3.$
Following the same steps, we can obtain a similar solution for $I_2$, i.\,e.\,, $I_2(t) = \bm{p}_2(t)^\intercal \bm{w} + q_2(t)g + c_4$, where we present $\bm{p}_2(t)$ and $q_2(t)$ in Eq.~(\ref{eq:dmp_p2_q2}).
\par\nobreak\vspace{-0.4cm}
{\small
\begin{align}
    \begin{aligned}
        \bm{p}_2(t) &= \frac{1}{\tau^2}\int_0^t \exp\Big(\frac{\alpha}{2\tau}t'\Big)x(t')\bm{\varphi}_x^\intercal \mathrm{d}t', \\
    q_2(t) &= \frac{\alpha}{2\tau} \bigg[\exp\Big(\frac{\alpha}{2\tau}t\Big)-1\bigg] \label{eq:dmp_p2_q2}   
    \end{aligned}
\end{align}
}%

% %%%%%%%%%%%%%%%%%%%%%%%%%%%%%%%%%%%%%%%%%%%%%%%%%%%%%%%%%%%%%%%%%%%%%%%%%%
% % IDMP
% %%%%%%%%%%%%%%%%%%%%%%%%%%%%%%%%%%%%%%%%%%%%%%%%%%%%%%%%%%%%%%%%%%%%%%%%%%
\subsection{\acrshortpl{dmp}' Linear Basis Functions Representation.}
\label{sbsec:linbear basis function}
Substituting the two integrals $I_1$ and $I_2$ in Eq.~(\ref{eq:dmp_closed_form_position}) by their derived form, the constants $c_3$ and $c_4$ can then be merged into $c_1$ and $c_2$, respectively. We can now express the position of \acrshortpl{dmp} in Eq.~(\ref{eq:dmp_closed_form_position}) as a summation of complementary functions $y_1$ and $y_2$, plus a linear basis function representation of the weights $\bm{w}$ and the goal attractor $g$
\par\nobreak\vspace{-0.4cm}
{\small
\begin{align}
    \begin{aligned}
    y &= c_1y_1 + c_2y_2 + \begin{bmatrix}
                            y_2\bm{p_2}-y_1\bm{p_1} & y_2q_2 - y_1q_1
                          \end{bmatrix}\begin{bmatrix}
                            \bm{w}\\g
                          \end{bmatrix} 
                          \\&\equiv~ c_1y_1 + c_2y_2 + \ipbt \bm{w}_g, \label{eq:idmp_pos}
    \end{aligned}                          
\end{align}
}%
% \begin{wrapfigure}{r}{0.5\textwidth}
\begin{figure}[t!]
    \vspace{0.2cm}
    \begin{center}
    \subcaptionbox{Weights' basis \label{fig:idmp_basis_pw}}[0.23\textwidth]{
    \includegraphics[width=0.23\textwidth]{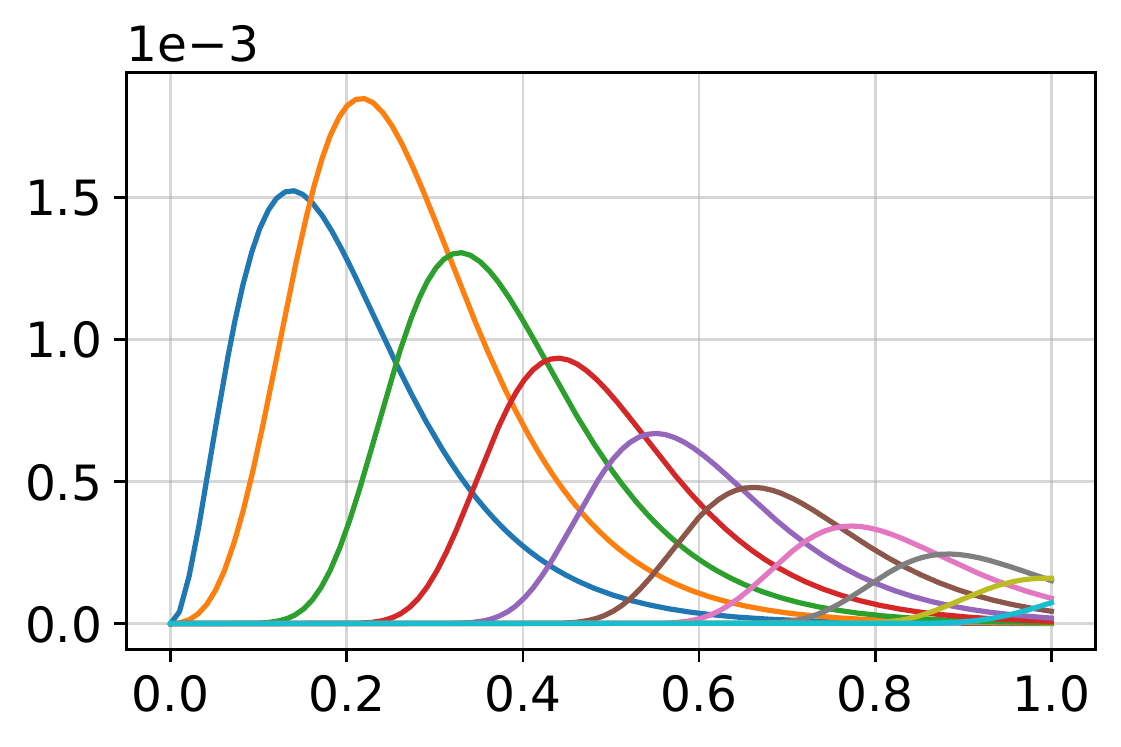}
  }
    \hfill
    \subcaptionbox{Goal's basis \label{fig:idmp_basis_pg}}[0.23\textwidth]{
    \includegraphics[width=0.23\textwidth]{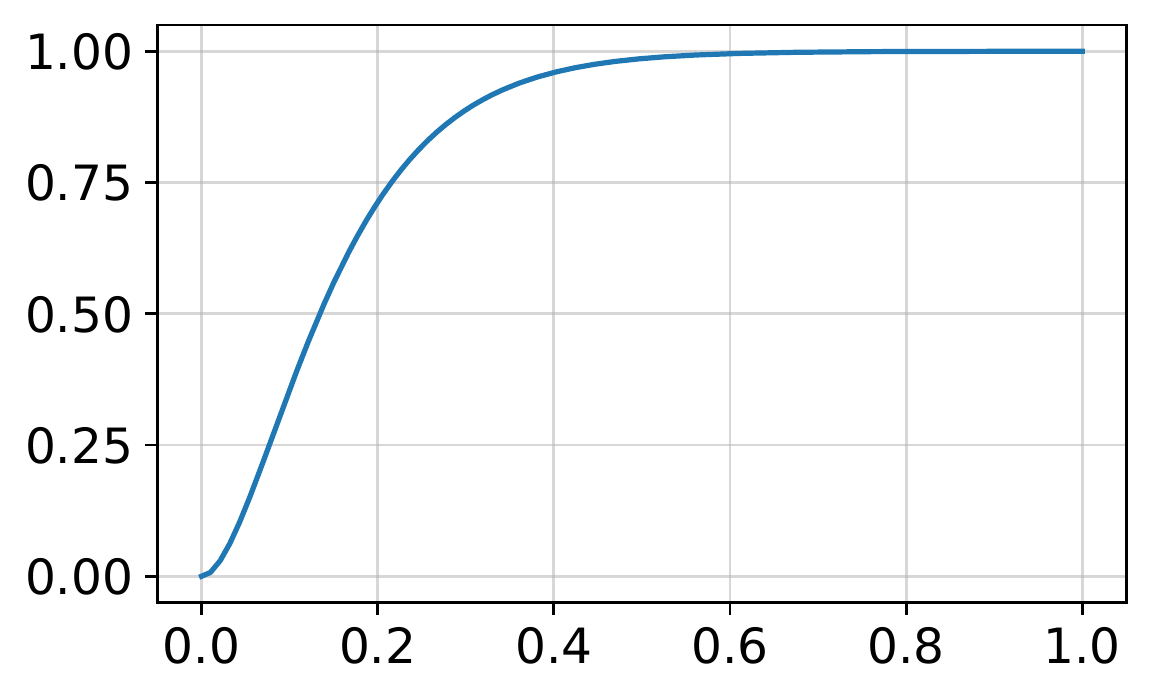}
  }
    \end{center}
    \vspace{-0.25cm}
    \caption{The basis functions $\ipb$ of the \acrshortpl{pdmp} can be computed offline and later used in online trajectory generation.}
    \label{fig:idmp_basis}
% \vspace{0.1cm}
\end{figure}
% \end{wrapfigure}

\def\sca{0.9}
\def\h{0.5cm}
\def\w{1.2cm}
\def\arrowblue{black!20!blue}
\def\arrowred{red}
\renewcommand{\arraystretch}{1.6}
\begin{figure}
    % \vspace{0.1cm}
    \centering
    \captionbox{Comparison of trajectory generation pipelines between (a) NN-based \acrshortpl{dmp} \cite{gams2018deep,ridge2020training,bahl2020neural} and (b) \acrshortpl{pdmp}. The node \textbf{DNN} represents arbitrary deep neural network architecture. The blue arrows denote the learning pipeline while the red arrows the numerical integration. Our method transforms the expensive numerical integration as basis functions computed offline which speeds up the trajectory computation and allows trajectory distribution prediction. \label{fig:workflow}}[\linewidth]{
        \subcaptionbox{NN + \acrshortpl{dmp} couple the numerical integration and the learning.\label{fig:nn_dmp_workflow}}[\linewidth]{
            \begin{tikzpicture}[
        	rrnode_data/.style={rectangle, minimum size=5mm, rounded corners=1mm, scale=\sca},
        	rrnode_nn/.style={rectangle, draw, minimum size=5mm, rounded corners=1mm, scale=\sca},
        	rrnode_ag/.style={circle, draw, minimum size=5mm, scale=\sca},
        	rrnode_rec/.style={circle, draw, minimum size=5mm, scale=\sca},
        	node_label/.style={circle, minimum size=4mm, scale=\sca}]]
        	
        	\begin{scope}[xshift=5cm,on grid]
        	    \node[rrnode_data]  (vel) at (0,0) {$\dot{y}$};
        	    \node[rrnode_data]  (acc)  [left = 0.85*\w of vel] {$\ddot{y}$};
    	        \node[rrnode_data]  (pos)  [right = 0.85*\w of vel] {$y$};
        	    \node[rrnode_data]  (pos_invisible)  [right = 0.8*\w of vel] {$\phantom{\dot{y}}$};
        	    
        	    \node[rrnode_data]  (f)  [left = 0.8*\w of acc] {$f$};
        	    \node[rrnode_data]  (wg)  [left = 0.9*\w of f] {$\bm{w}_g$};
        	    \node[rrnode_nn]    (dnn) [left = 1.2*\w of wg]  {\textbf{DNN}};
    	        \node[rrnode_data]    (online) [above right = 1.2*\h and 0.04*\w of dnn]  {\textbf{Online:}};
    	        \node[rrnode_data]  (mse)  [right = 1.3*\w of pos] {MSE Loss};
    	        
    	        \draw[-{Stealth[scale=0.6]}, ultra thick, color=\arrowred] (pos_invisible.north) -- +(0, 0.9em) -| (acc.north);
    	        \draw[-{Stealth[scale=0.6]}, ultra thick, color=\arrowred] (vel.north) -- +(0, 0.9em) -| (acc.north);
    	        
    	        \draw[-{Stealth[scale=0.8]}, ultra thick, color=\arrowblue] (dnn.east)+(0.3em, 0em) -- (wg);
    	        \draw[-{Stealth[scale=0.8]}, ultra thick, color=\arrowblue] (wg) -- (f);
    	        \draw[-{Stealth[scale=0.8]}, ultra thick, color=\arrowblue] (f) -- (acc);
    	        
    	        \draw[-{Stealth[scale=0.8]}, ultra thick, color=\arrowred] (acc) -- (vel);
    	        \draw[-{Stealth[scale=0.8]}, ultra thick, color=\arrowred] (vel) -- (pos);
    	        
    	        \draw[-{Stealth[scale=0.8]}, ultra thick, color=\arrowblue] (pos) -- (mse);
   	        \end{scope}
        \end{tikzpicture}
        \vspace{-0.3cm}
        \hspace{0.9cm}
        }
        \subcaptionbox{NN + \acrshortpl{pdmp} decouple the numerical integration from the learning pipeline and can also model trajectory distributions.
        \label{fig:nn_idmp_workflow}}[\linewidth]{
            \vspace{0.3cm}
            \begin{tikzpicture}[
        	rrnode_data/.style={rectangle, minimum size=5mm, rounded corners=1mm, scale=\sca},
        	rrnode_nn/.style={rectangle, draw, minimum size=5mm, rounded corners=1mm, scale=\sca},
        	rrnode_ag/.style={circle, draw, minimum size=5mm, scale=\sca},
        	rrnode_rec/.style={circle, draw, minimum size=5mm, scale=\sca},
        	node_label/.style={circle, minimum size=4mm, scale=\sca}]]
        	
        	\begin{scope}[xshift=0cm,on grid]
        	
        	    \node[rrnode_nn]    (dnn) at (0,0)  {\textbf{DNN}};
        	    \node[rrnode_data]    (online) [above right = 1.2*\h and 0.04*\w of dnn]  {\textbf{Online:}};
        	    \node[rrnode_data]    (offline) [above = 1.3*\h of online]  {\textbf{Offline,}};
        	    \node[rrnode_data]    (once) [right = 1.45*\w of offline]  {\textbf{computed once:}};
        	    \node[rrnode_data]  (wg)  [right = 2*\w of dnn, text width=1cm] {$\bm{w}_g$ or $p(\bm{w}_g)$};
        	    
        	    \node[rrnode_data]  (merge)  [right = 1.2*\w of wg] {$\phantom{.}$};
    	        
    	        \node[rrnode_data]  (basis)  [above = 2.5*\h of merge] {$\ipb$};
    	        \node[rrnode_data]  (basis_invisible)  [left = 0.8*\w of basis] {$\phantom{\ipb}$};
    	        
    	        \node[rrnode_data]  (pos)  [right = 2.2*\w of wg, text width=1cm] {$\bm{\lambda}$ or $p(\bm{\lambda})$};
    	        
    	        \node[rrnode_data]  (mse)  [right = 2*\w of pos, text width=1.5cm] {MSE or LL Loss};
    	        
    	        \draw[color=black!20!white, thick] (-0.5, 0.9) -- (7.8,0.9);
    	        
    	        \draw[-{Stealth[scale=0.8]}, ultra thick, color=\arrowblue] (dnn.east)+(0.3em, 0em) -- (wg);
    	        \draw[-{Stealth[scale=0.8]}, ultra thick, color=\arrowblue] (wg) -- (pos);
    	        \draw[-{Stealth[scale=0.8]}, ultra thick, color=\arrowblue] (pos) -- (mse);
    	        \draw[-{Stealth[scale=0.8]}, ultra thick, color=\arrowblue] (basis) |- (pos);
    	        \draw[-{Stealth[scale=0.8]}, ultra thick, color=\arrowred] (basis_invisible) -- (basis);
      	    \end{scope}
        \end{tikzpicture}
        }
    }
    \vspace{-0.3cm}
\end{figure}
\begin{table}[t!]
    \centering
        \captionof{table}{Computation time of both pipelines. Here a 2-DoFs, 6-seconds long, 1000 Hz trajectory is generated from a 22-dim $\bm{w}_g$ parameter vector. We tested both forward pass (\textbf{FP}) and backward pass (\textbf{BP}). A 3-layer fully connected (FC) network with [10, 128, 22] neurons on input, hidden and output layers respectively are used to simulate the learning procedure. The keyword \textbf{+BC} are the settings where the boundary conditions are renewed so that the coefficients $c_1$ and $c_2$ need to be recomputed. Otherwise, they remain unchanged. The result shows that our model is 200-4600 times faster than the NN-\acrshortpl{dmp} in different settings. We use a Nvidia\textregistered ~RTX-3080Ti GPU for our test. In a full learning experiment with NN architectures, this speed difference translates into a speed-up of around 10 times (see experiments). 
        % reducing learning time from 105 min to 10 min.
        \label{table:speed_test}}{
        \scriptsize
        \begin{tabular}{ccccc}
            \toprule
            \textbf{Pipelines} &\textbf{FP}&\textbf{FP + BC} & \textbf{BP} & \textbf{BP + BC} \\
            \hline
            \textbf{NN-\acrshortpl{dmp}} & 0.6057 s & 0.6145 s & 1.5261 s & 1.5737 s\\
            \textbf{\acrshortpl{pdmp}}& 0.00013 s & 0.0027 s & 0.00105 s & 0.0039 s \\
            \textbf{Speed-up} & \textbf{$\bf{\times}$ 4659} & \textbf{$\bf{\times}$ 227} & \textbf{$\bf{\times}$ 1453} & \textbf{$\bf{\times}$ 403}\\
            \bottomrule
        \end{tabular}
    }
% \vspace{-0.2cm}
\end{table}
\renewcommand{\arraystretch}{1.5}

\noindent where $\bm{w}_g$ is a N+1-dim vector containing the weights $\bm{w}$ and the goal $g$. 
The resulting basis functions for $\bm{w}$ and $g$ are represented by $\ipb(t)$, which can be solved numerically, cf. Fig.~\ref{fig:idmp_basis}. 
The constants $c_1$ and $c_2$ are determined by solving a \acrlong{bc} problem where we use the current position and velocity of the robot to inscribe where the trajectory should start.

In contrast to previous NN-\acrshortpl{dmp} methods \cite{gams2018deep,ridge2020training,bahl2020neural}, our model separates the learnable parameters from the numerical integrals which are transformed as basis functions.
These basis functions are shared by all trajectories to be generated during learning procedure. Hence, we can pre-compute these basis functions once offline at first and use them as constants in online trajectory generation. Consequently, we exclude numerical integrals from forward and backward propagation of NN pipelines, and significantly increase the trajectory generation speed.
We present a pipeline comparison and corresponding speed test between previous methods and our \acrshortpl{pdmp} in Fig.~\ref{fig:workflow} and TABLE~\ref{table:speed_test}. 
%\reb{Further, we also compare the computation time in our experiment.}
% \vspace{-0.4cm}
%%%%%%%%%%%%%%%%%%%%%%%%%%%%%%%%%%%%%%%%%%%%%%%%%%%%%%%%%%%%%%%%%%%
%  Solve BC
%%%%%%%%%%%%%%%%%%%%%%%%%%%%%%%%%%%%%%%%%%%%%%%%%%%%%%%%%%%%%%%%%%%
\subsection{Solving the Boundary Condition Problem}
\label{sbsec:idmp_bc}
To compute the coefficients $c_1$ and $c_2$, we need to first obtain the velocity representation by computing the derivative of Eq.~(\ref{eq:dmp_closed_form_position}) w.r.t. time. The resulting equation reads
\par\nobreak\vspace{-0.1cm}
{\small
\begin{equation}
    \dot{y}=c_1\dot{y}_1 + c_2\dot{y}_2 - \dot{y}_1\int\frac{y_2 F}{Y}\mathrm{d}t + \dot{y}_2\int\frac{y_1 F}{Y}\mathrm{d}t.
    \label{eq:dmp_closed_form_velocity}
\end{equation}
}%
Note that Eq.~(\ref{eq:dmp_closed_form_position}) and Eq.~(\ref{eq:dmp_closed_form_velocity}) share a similar structure using the same constants $c_1$ and $c_2$, as well as the two indefinite integrals in Eq.~(\ref{eq:dmp_i1_i2}). The only difference is that the two complementary functions $y_1$ and $y_2$ are replaced by their derivatives $\dot{y}_1$ and $\dot{y}_2$.
By reusing the derivation of Eq.~(\ref{eq:dmp_i1_i2}, \ref{eq:dmp_i1_decouple} ,\ref{eq:dmp_p1_q1}, \ref{eq:dmp_p2_q2}), the trajectory velocity is ultimately represented by 
\par\nobreak\vspace{-0.5cm}
{\small
\begin{align}
    \begin{aligned}
        \dot{y} &= c_1\dot{y}_1 + c_2\dot{y}_2 + \begin{bmatrix}
                                \dot{y}_2\bm{p_2}-\dot{y}_1\bm{p_1} \dot{y}_2q_2 - \dot{y}_1q_1
                            \end{bmatrix}\begin{bmatrix}\vspace{-0.2cm}\bm{w}\\g
                            \end{bmatrix}\\
    &\equiv c_1\dot{y}_1 + c_2\dot{y}_2 + \ivbt\bm{w}_g,
     \label{eq:idmp_vel}
    \end{aligned}
\end{align}
}%
where $\ivb(t)$ represents the basis functions of the velocity.
The position and velocity in Eq.~(\ref{eq:idmp_pos}, \ref{eq:idmp_vel}) share the same linear model structure, coefficients $c_1, c_2$, and the parameters $\bm{w}_g$. 

We need two boundary conditions to solve $c_1$ and $c_2$. Theoretically, there are three options, \begin{enumerate*}[label=(\alph*)]
    \item two boundary position values at two time steps, i.\,e.\, $y(t_{b_1}),~y(t_{b_2}),~t_{b_1} \neq t_{b_2}$,
    \item two boundary velocity values at two time steps, i.\,e.\, $\dot{y}(t_{b_1}),~\dot{y}(t_{b_2}),~t_{b_1} \neq t_{b_2}$, and
    \item one boundary position value plus one boundary velocity value $y(t_{b_1}),~\dot{y}(t_{b_2})$, where $t_{b_1}$ and $t_{b_2}$ are typically identical. 
\end{enumerate*} 
The third option allows us to naturally integrate the current position and velocity of a robot as boundary conditions of a trajectory to be generated.
% Consequently, our representation guarantees that the planned trajectory always satisfies any pair of position and velocity at a certain time step, if used as boundary conditions.
% Such boundary conditions do not have to always be the values at the initial time step $t_0~(=0)$ but can be specified for any replanning time step.
% To make the derivation clear, we recall the position Eq.~(\ref{eq:idmp_pos}) and velocity Eq.~(\ref{eq:idmp_vel}) representations of \acrshortpl{pdmp} at one single time step, i.e.,
% \par\nobreak\vspace{-0.2cm}
% {\small
% \begin{equation*}
%     y = c_1y_1 + c_2y_2 + \ipbt\bm{w}_g, \quad\dot{y} = c_1\dot{y}_1 + c_2\dot{y}_2 + \ivbt\bm{w}_g.
% \end{equation*}
% }%
We denote the boundary conditions as a position and velocity pair ($y_b, \dot{y}_b$) at time step $t_b$. The values of the complementary functions in Eq.~(\ref{eq:dmp_y1_y2}) and their corresponding derivatives at $t_b$ are denoted by $y_{1_b}, y_{2_b}, \dot{y}_{1_b} \dot{y}_{2_b}$. The value of the position basis and velocity basis at $t_b$ are denoted as $\ipbb, \ivbb$. 
By substituting these terms into Eq.~(\ref{eq:idmp_pos}, \ref{eq:idmp_vel}), we can solve
\par\nobreak\vspace{-0.3cm}
{
\begin{equation}
    {\small
    \begin{bmatrix}c_1\\c_2\end{bmatrix}
    }%
    = \begin{bmatrix}\frac{\dot{y}_{2_b}y_b-y_{2_b}\dot{y}_b}{y_{1_b}\dot{y}_{2_b}-y_{2_b}\dot{y}_{1_b}} +\frac{y_{2_b}\ivbbt-\dot{y}_{2_b}\ipbbt}{y_{1_b}\dot{y}_{2_b}-y_{2_b}\dot{y}_{1_b}}\bm{w}_g\\[0.7em]
    \frac{y_{1_b}\dot{y}_b-\dot{y}_{1_b}y_b}{y_{1_b}\dot{y}_{2_b}-y_{2_b}\dot{y}_{1_b}}+\frac{\dot{y}_{1_b}\ipbbt- y_{1_b}\ivbbt}{y_{1_b}\dot{y}_{2_b}-y_{2_b}\dot{y}_{1_b}}\bm{w}_g\end{bmatrix}.
    \label{eq:dmp_c}
\end{equation}
}%
Substituting Eq.~(\ref{eq:dmp_c}) into Eq.~(\ref{eq:idmp_pos}), we can express the trajectory position as
\par\nobreak\vspace{-0.1cm}
{\small
\begin{equation}
    y= \xi_1 y_b  + \xi_2 \dot{y}_b +[\xi_3 \ipbb + \xi_4 \ivbb+ \ipb]^\intercal\bm{w}_g, \label{eq:idmp_pos_full}
\end{equation}
}%
where 
\vspace{-0.5cm}
\par\nobreak
{\footnotesize
\begin{align*}
    \xi_1 &= \xi_1(t)=\frac{\dot{y}_{2_b}y_1-\dot{y}_{1_b}y_2}{y_{1_b}\dot{y}_{2_b}-y_{2_b}\dot{y}_{1_b}},\quad\quad\xi_2 =\xi_2(t)=\frac{y_{1_b}y_2-y_{2_b}y_1}{y_{1_b}\dot{y}_{2_b}-y_{2_b}\dot{y}_{1_b}},\\
    \xi_3 &= \xi_3(t)=\frac{\dot{y}_{1_b}y_2-\dot{y}_{2_b}y_1}{y_{1_b}\dot{y}_{2_b}-y_{2_b}\dot{y}_{1_b}},\quad\quad\xi_4 =\xi_4(t)=\frac{y_{2_b}y_1- y_{1_b}y_2}{y_{1_b}\dot{y}_{2_b}-y_{2_b}\dot{y}_{1_b}}.
\end{align*}
}%
% Replacing $y_1$ and $y_2$ in $\xi_k,~k\in\{1,2,3,4\}$ by $\dot{y}_1$ and $\dot{y}_2$, respectively, yields $\dot{\xi}_k$.
Eq.~(\ref{eq:idmp_pos_full}) shows that the trajectory position $y$ is fully determined by the boundary position $y_b$, and velocity $\dot{y}_b$ at the time step $t_b$, as well as the learnable weights $\bm{w}_g$.

\subsection{Probability Distribution of Multi \acrshortpl{dof} \acrshortpl{dmp}}
\begin{figure*}[t!]
\centering
\vspace{0.2cm}
% dmp vs prodmp
\subcaptionbox{Fit BC in reproduction\label{fig:dmp_vs_promp_vs_prodmp}}
[.247\textwidth]{
    \includegraphics[width=\linewidth]{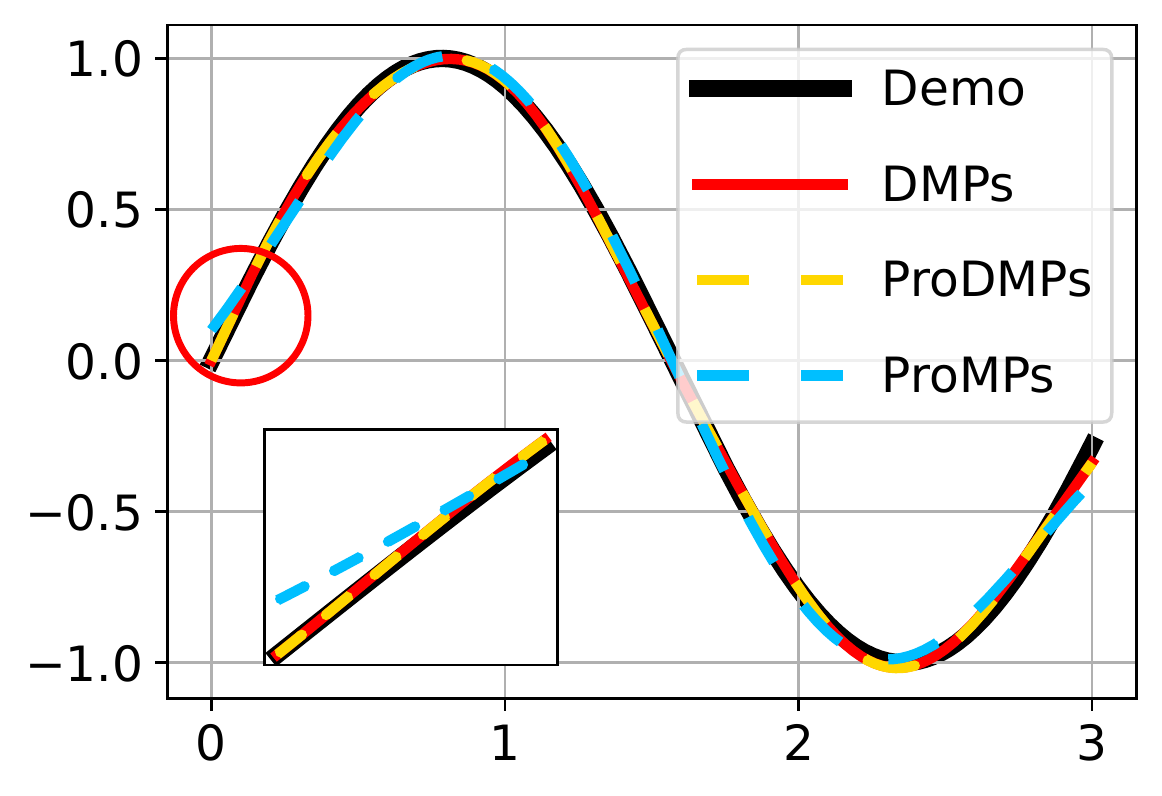}
    }%
\hfill
% \hspace{0.3mm}
% prodmp pos
\subcaptionbox{Fit BC in distribution\label{fig:dmp_vs_prodmp}}
[.249\textwidth]{
    \includegraphics[width=\linewidth]{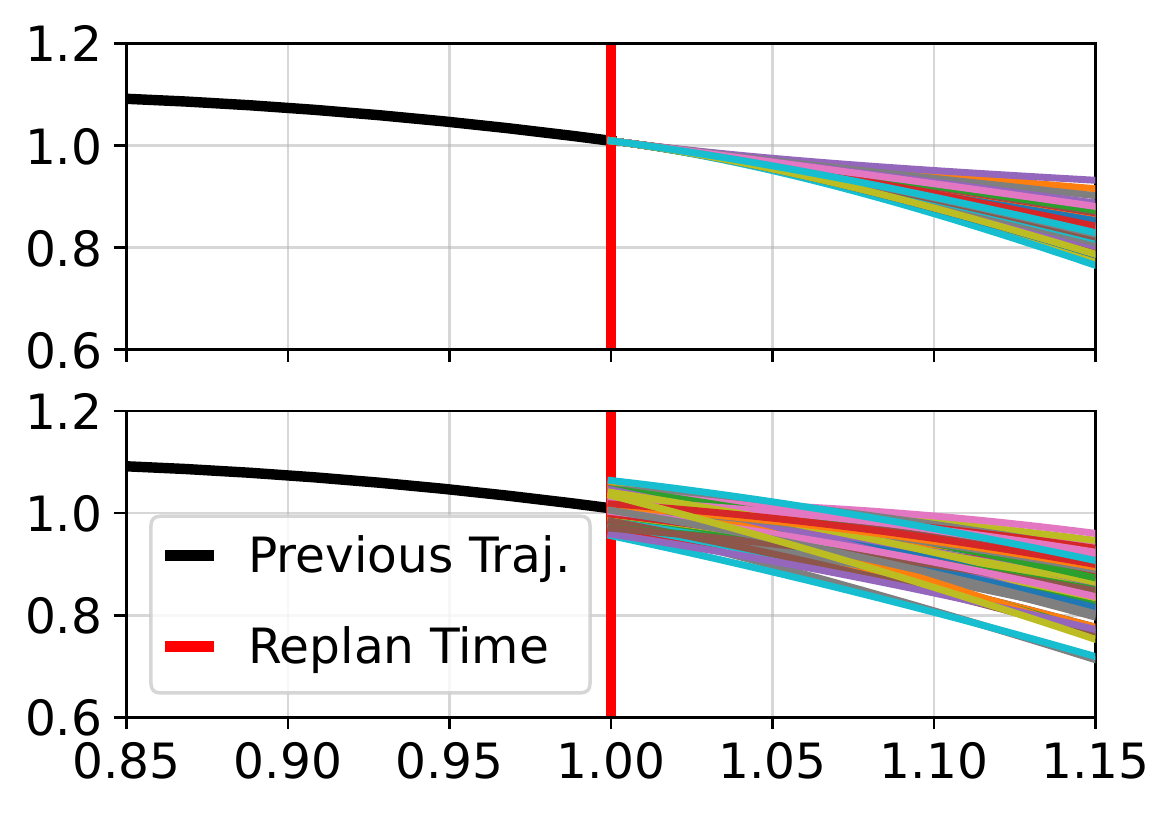}
    }%
\hfill
% combination
\subcaptionbox{Combination\label{fig:combination}}
[.247\textwidth]{
    \includegraphics[width=\linewidth]{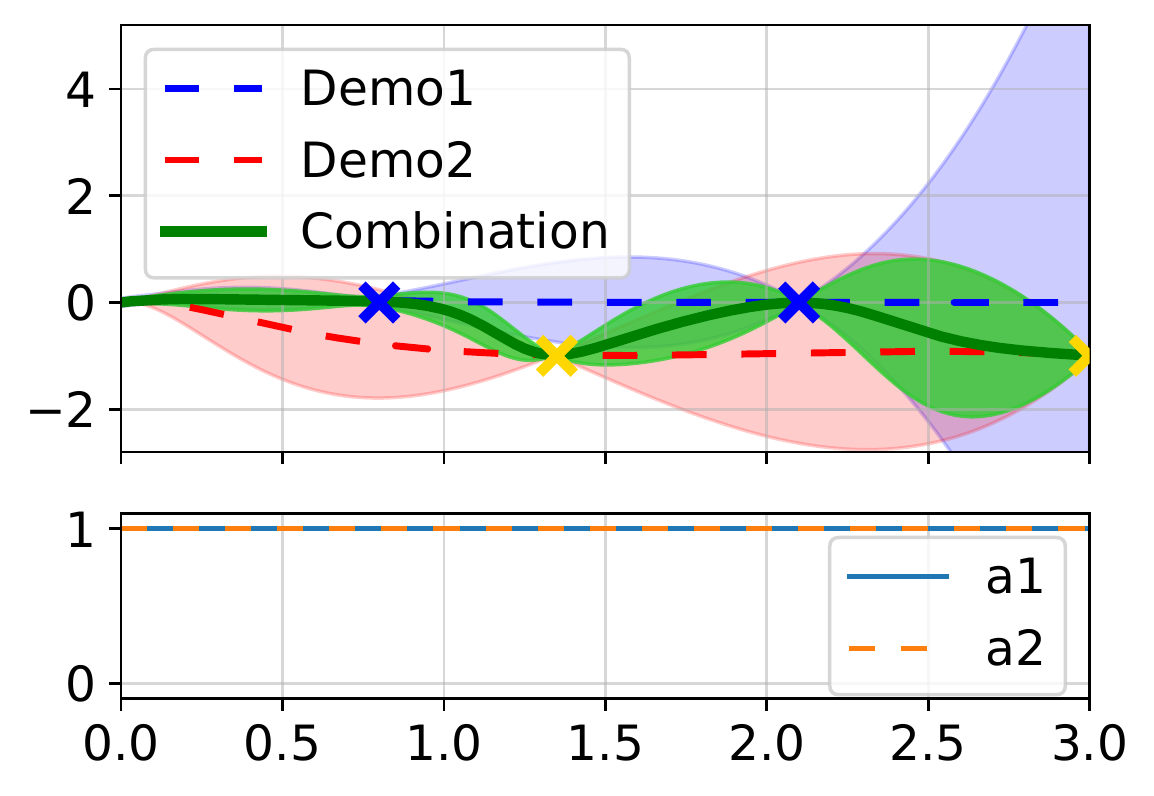}
    }%
\hfill
% blending
\subcaptionbox{Blending\label{fig:blending}}
[.247\textwidth]{
    \includegraphics[width=\linewidth]{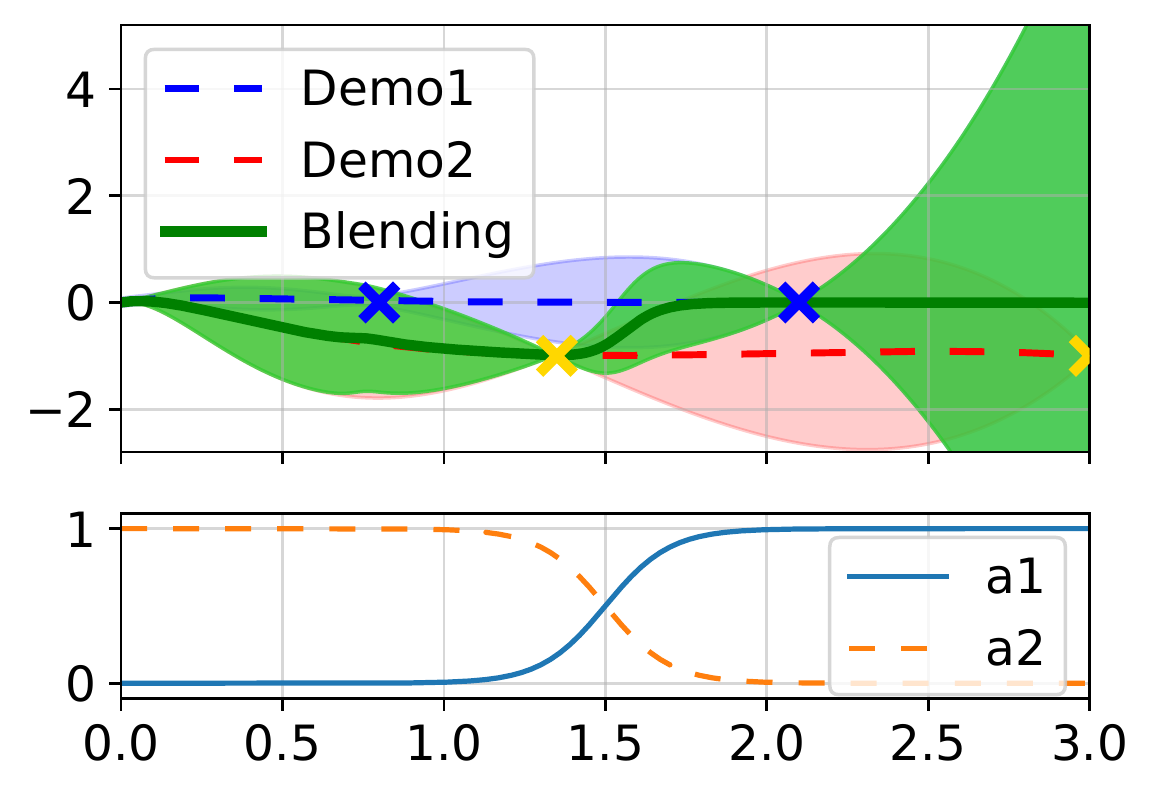}
    }%
\caption{Illustration of \acrshortpl{pdmp}' properties inherited from \acrshortpl{dmp} and \acrshortpl{promp}. (a) Given a demonstrated trajectory, \acrshortpl{pdmp} can reproduce it and mathematically guarantee a fit of the boundary condition (BC), which cannot be guaranteed by \acrshortpl{promp}. Besides, \acrshortpl{pdmp} reproduce the identical trajectory of \acrshortpl{dmp}', given the same parameters $\bm{w}_g$. As our method uses the linear basis functions rather than online numerical integration, it is thus much faster. (b) \acrshortpl{pdmp} (upper) can model trajectory distributions and guarantee that all newly sampled trajectories follow the given boundary conditions, i.e. smooth online replanning. In the contrast, \acrshortpl{promp}' samples (lower) cannot fit the boundary condition well, which often results in a jump in the case of replanning. (c) + (d) \acrshortpl{pdmp} support probabilistic operations of \acrshortpl{promp}, e.g. combination and blending \cite{paraschos2013probabilistic} of two demonstrated trajectory distributions (blue and red) into one resulting distribution (green).}
\label{fig:prodmp_property}
\vspace{-0.1cm}
\end{figure*}
The linear basis function representation of \acrshortpl{pdmp} takes the same form of \acrshortpl{promp}. Hence, our model can also be extended to multi-\acrshortpl{dof} systems $\bm{y}=[y^1,~...,~y^D]^\intercal$, where $D$ denotes the system's \acrshortpl{dof}. 
Similar to \cite{paraschos2013probabilistic}, we extend the basis functions $\ipb, \ivb$, as well as their boundary values $\ipbb, \ivbb$ to block-diagonal matrices $\bipb, \bivb$, $\bipbb, \bivbb$ and concatenate each \acrshort{dof}'s boundary conditions into a vector, e.\,g.\,, $y^1_b,~...,~y^D_b ~\rightarrow \bm{y}_b = [y^1_b,~...,~y^D_b]^\intercal$. Additionally, we note that the coefficient constants $c_1$ and $c_2$ of each \acrshort{dof} can be solved independently. As a consequence, we have the multi-\acrshortpl{dof} trajectory as
\par\nobreak\vspace{-0.1cm}
{\small
\begin{equation}
    \bm{y} = \xi_1  \bm{y}_b + \xi_2  \dot{\bm{y}}_b +[\xi_3  \bipbb + \xi_4  \bivbb+ \bipb]^\intercal\bm{w}_g,\label{eq:idmp_pos_multi_full}
\end{equation}
}%
% The terms $\xi_k, \dot{\xi}_k,~k\in\{1,2,3,4\}$ remain the same definition in the single \acrshort{dof}, as they are independent to the number of \acrshortpl{dof}.

%%%%%%%%%%%%%%%%%%%%%%%%%%%%%%%%%%%%%%%%%%%%%%%%%%%%%%%%%%%%%%%%%%%%%%%%%%
% idmp_uncertainty
%%%%%%%%%%%%%%%%%%%%%%%%%%%%%%%%%%%%%%%%%%%%%%%%%%%%%%%%%%%%%%%%%%%%%%%%%

\label{app:idmp_promp}
Extending Eq.~(\ref{eq:idmp_pos_multi_full}) from a single time step to the entire trajectory $\bm{\Lambda} = [\bm{y}_t]_{t=0:T}$, we can now express the trajectory distribution similarly to \acrshortpl{promp}.
We consider a system where the current robot state $\bm{y}_b, \dot{\bm{y}}_b$ can be acquired precisely, which is universal in most robotic systems. 
In this case, the trajectory's variability is only caused by the variability of the weights $\bm{w}_g$ plus the observation white noise $\bm{\epsilon}_y$.
Similar to \acrshortpl{promp}, assuming the weights $\bm{w}_g$ follows a multivariate normal distribution $\bm{w}_g\sim\mathcal{N}(\bm{w}_g|\bm{\mu_{w_g}},\bm{\Sigma_{w_g}})$, we can compute the trajectory distribution with full covariance over all time steps $0:T$ and all \acrshortpl{dof} $1:D$ as
\par\nobreak\vspace{-0.2cm}
{\small
\begin{equation}
    p(\bm{\Lambda};\bm{\mu}_{w_g},\bm{\Sigma}_{w_g}, \bm{y}_b, \dot{\bm{y}}_b) = ~\mathcal{N}(\bm{\Lambda}|~\bm{\mu}_{\Lambda}, \bm{\Sigma}_{\Lambda}),\label{eq:idmp_mean_cov}
\end{equation}
\vspace{-0.6cm}
\begin{align*}
    \begin{aligned}
    \bm{\mu}_{\Lambda} &= \bm{\xi}_1 \bm{y}_b + \bm{\xi}_2 \dot{\bm{y}}_b + \bm{H}^\intercal_{0:T} \bm{\mu}_{w_g}, \\
    \bm{\Sigma}_{\Lambda} &=\bm{H}^\intercal_{0:T}\bm{\Sigma}_{w_g} \bm{H}_{0:T}+ \sigma_n^2 \mathbf{I},\\
    \text{with}~~\bm{H}_{0:T} &= \bm{\xi}_3 \bipbb + \bm{\xi}_4 \bivbb+ \bipb_{0:T}
    \end{aligned}
\end{align*}
}%
where $\bm{\xi}_k = [\xi_k(t)]_{t=0:T}$.
% The dimensions of all variables in Eq.~(\ref{eq:idmp_mean_cov}) are described in Table \ref{tab:dim_variables}.
% \input{table/input_output}
In a summary, \acrshortpl{pdmp} unify \acrshortpl{dmp} and \acrshortpl{promp} in a consistent framework, inheriting the properties and advantages of both methodologies.
In Fig.~\ref{fig:prodmp_property}, we illustrate a few examples to highlight the properties. We show that \acrshortpl{pdmp} using linear basis functions can generate an identical trajectory as \acrshortpl{dmp}, and enforce the predicted trajectory distribution adheres the given boundary conditions. The newly generated trajectories are guaranteed to have a smooth transition to the previous trajectory at the replanning time step. Furthermore, \acrshortpl{pdmp} inherit all the probabilistic operations of \acrshortpl{promp}, such as combination and blending. For the exact formulations of these operations, we refer to the original \acrshortpl{promp} paper \cite{paraschos2013probabilistic}, as they remain unaltered.

\section{Embed \acrshortpl{pdmp} in a deep architecture using Bayesian Set Encoders}
\label{sec:nmp}
We extend \acrshortpl{pdmp} with a deep NN architecture that allows for conditioning the trajectory distribution on a set of high-dimensional (time-stamped) sensory observations, such as images. Similar to recent \acrshort{mp} architectures \cite{seker2019conditional, akbulut2021acnmp}, we treat such observations as set-inputs \cite{zaheer2017deep, garnelo2018neural}, i.\,e.\,, our architecture is invariant to the order of these observations. 

%%%%%%%%%%%%%%%%%%%%%%%%%%%%%%%%%%%%%%%%%%%%%%%%%%%%%%%%%%%%%%%%%%%%%%%%%%
% Architecture
%%%%%%%%%%%%%%%%%%%%%%%%%%%%%%%%%%%%%%%%%%%%%%%%%%%%%%%%%%%%%%%%%%%%%%%%%%
\textbf{Architecture.}
\label{sbsec:nmp_model}
The architecture contains four major parts: encoder, aggregator, decoder, and the \acrshortpl{pdmp} layer, cf. Fig.~\ref{fig:nmp_structure}. 
The \emph{Encoders} ${E}_{\bm{\mu}}$ and ${E}_{\bm{\sigma}^2}$ of each sensor input type, e.\,g.\, images, compute for each observation $\bm{o}_{m} \in \mathcal{O}$ a corresponding latent observation $\obslat$ and an uncertainty $\obsunc$, measuring how informative this observation is. Different types of sensor inputs, e.\,g.\,, images and robot states, are mapped via different types of encoders, e.\,g.\,, CNN encoders and MLP encoders, into the same latent space.
The \emph{\acrlong{ba}} \cite{volpp2020bayesian} ${A}$ is a parameter-free operator used to aggregate a set of latent observations $\{\obslat\}$ and the corresponding uncertainties $\{\obsunc\}$ into a latent state posterior $p(\bm{z}|\mathcal{O})$ which is given by a factorized multivariate Gaussian distribution $\mathcal{N}(\bm{z}|\bm{\mu_z}, \bm{\sigma_z}^2)$. The resulting aggregation is
\par\nobreak\vspace{-0.35cm}
{\footnotesize
\begin{align}
    \begin{aligned}
    \bm{\sigma}_{\bm{z}|\bm{o}_{1:M}}^2 &= \left[(\zpriorvar)^{\ominus} + \sum_{m=1}^M(\obsunc)^{\ominus}\right]^{\ominus}, \\
    \bm{\mu}_{\bm{z}|\bm{o}_{1:M}} &= \zpriormu + \bm{\sigma_z}^2 \odot \sum_{m=1}^M(\obslat - \zpriormu)\oslash \obsunc,\label{eq:ba}
    \end{aligned}
\end{align}
}%
where $\ominus$, $\odot$, and $\oslash$ denote element-wise inversion, product, and division, respectively. 
The parameters of the latent variable's prior distribution are $\zpriormu$ and $\zpriorvar$. Intuitively, \acrshort{ba} takes the uncertainty of the different conditioning events into account which typically leads to better results than the mean aggregation \cite{volpp2020bayesian}. Hence, we adopt this approach for our architecture.
The \emph{Decoders} $D_{\mu}, D_L$ predict the mean $\bm{\mu_{w_g}}$ and the Cholesky decomposition $\bm{L_{w_g}}$ of the covariance $\bm{\Sigma_{w_g}}$ of the weights distribution. Similar to \citet{volpp2020bayesian}, we do not sample from the latent posterior $p(\bm{z}|\mathcal{O})$ but use its mean $\bm{\mu}_z$ and variance $\bm{\sigma}^2_z$ to predict $\bm{\mu_{w_g}}$ and $\bm{L_{w_g}}$, respectively.
Given the query time steps $\{t\}$ and the current robot position $\bm{y}_b$ and velocity $\dot{\bm{y}}_b$ as boundary conditions, the \emph{\acrshortpl{pdmp}} layer computes the parameters $\bm{\mu_{\Lambda}}$ and $\bm{\Sigma_{\Lambda}}$ of the trajectory distribution as a multivariate Gaussian distribution. 

% %%%%%%%%%%%%%%%%%%%%%%%%%%%%%%%%%%%%%%%%%%%%%%%%%%%%%%%%%%%%%%%%%%%%%%%%%%
% % Learning objective
% %%%%%%%%%%%%%%%%%%%%%%%%%%%%%%%%%%%%%%%%%%%%%%%%%%%%%%%%%%%%%%%%%%%%%%%%%%
\textbf{Loss Function.}
% \label{sbsec:nmp_obj}
In this paper, we focus on \acrlong{il} tasks and minimize the negative log-likelihood of the conditional trajectory distribution, i.\,e.\,, $-\log\mathcal{N}(\bm{\Lambda}|\bm{\mu}_{\Lambda}, \bm{\Sigma}_{\Lambda})$. Here, $\bm{\Lambda}=\{\bm{y}_t\}_{t=0...T}$ is the trajectory ground truth, and $\bm{\mu}_{\Lambda}, \bm{\Sigma}_{\Lambda}$ the mean and covariance of the predicted trajectory distribution conditioned on a set of observations $\mathcal{O}$. 
Yet, predicting a trajectory's full covariance matrix $\bm{\Sigma}_{\Lambda}$ demands high computational resources. 
The $TD \times TD$ covariance matrix
% ($T$: number of time steps, $D$: number of \acrshortpl{dof})~
has to be inverted in the loss computation.
% \cite{bientinesi2008families}.  
To keep the computation manageable, we never compute the distribution of the whole time sequence, but only compute the covariance of a pair of time steps, which limits the size of covariance matrices to $2D \times 2D$. We randomly select $J$ such time pairs $\{(t, t')_j\}_{j=1,...,J}$, where $t, t' \in \{0,..., T\}$, and predict the joint distribution on each of these pairs. 
As such random selection is executed in every training batch, we can still learn the correlation between time steps while keeping the cost of matrix inversion manageable.
Given the \acrlongpl{bc} $\bm{y}_b, \dot{\bm{y}}_b$ and a set of observations $\mathcal{O}$, the loss function is thus defined as the mean negative log-likelhood of $J$ random pairs of trajectory values $\bm{y}_{(t, t')_j}$ as
\par\nobreak\vspace{-0.4cm}
{\small
\begin{equation}
    % \text{Paired-Time (PT) loss:}\quad
    \mathcal{L}_{\bm{\theta}}(\bm{\Lambda}, \bm{y}_b, \dot{\bm{y}}_b, \mathcal{O}) = -\frac{1}{J}\sum_{j=1}^J\log \mathcal{N}(\bm{y}_{(t, t')_j} | \bm{\mu}_{(t, t')_j}, \bm{\Sigma}_{(t, t')_j}),
    \label{eq:rec_ll}
    % \vspace{-0.03cm}
\end{equation}
}%

% geometry constants
\def\sca{0.7}
\def\h{0.5cm}
\def\w{1.4cm}
\def\arrowblue{black!20!blue}
\def\arrowred{red}
\def\scale_factor{1.2}
\begin{figure}[t!]
    % \vspace{-0.5cm}
    \captionbox{Architecture and pipeline of our deep \acrshortpl{pdmp} model. The encoders, aggregator and decoders are denoted by $\bm{E}, \bm{A}$, and $\bm{D}$ respectively. Yellow nodes denote deep neural networks, while gray nodes denote operations without learnable parameters. \label{fig:nmp_structure}}[0.48\textwidth]{
        \begin{tikzpicture}[scale=0.8,
    	rrnode_data/.style={rectangle, minimum size=5mm, rounded corners=1mm, scale=\sca},
    	rrnode_box/.style={rectangle, draw=PineGreen, minimum width=0.7*\w, minimum height=2.5*\h, scale=\sca, fill=Goldenrod, opacity=0.4, text opacity=1},
    	rrnode_nn/.style={rectangle, draw, minimum size=5mm, rounded corners=1mm, scale=\sca},
    	rrnode_ag/.style={rectangle, draw, minimum width=0.5*\w, minimum height=2.5*\h, rounded corners=1mm, scale=\sca, draw=Gray, fill=Gray, fill opacity=0.3, text opacity=1}]

    	\begin{scope}[xshift=5cm,on grid]
    	    \node[rrnode_data]  (ctx) at (0,3)  {$\{\bm{o}_{m}\}$};
	        
	        \node[rrnode_box]    (encoder_mean)  [above right = \h and 1*\w of ctx] {$\bm{E_{\mu}}$};
	        \node[rrnode_box]    (encoder_var)   [below right = \h and 1*\w of ctx] {$\bm{E_{\sigma^2}}$};

	        \node[rrnode_ag]    (aggregator)    [below right = \h and 1.1*\w of encoder_mean] {$\bm{A}\,$};
 	        
 	        % Values around aggregator
 	        \node[rrnode_data]  (r_mean)    [above left = 1.2*\h and 0.5*\w of aggregator] {$\{\bm{r_m}\}$}; 
	        \node[rrnode_data]  (r_var)     [below left = 1.3*\h and 0.5*\w of aggregator] {$\{\bm{\sigma^2_{r_m}}\}$ };
	        \node[rrnode_data]  (z_mean)    [above right = 1.2*\h and 0.5*\w of aggregator] {$\bm{\mu_z}$};
	        \node[rrnode_data]  (z_var)     [below right = 1.3*\h and 0.5*\w of aggregator]{$\bm{\sigma_z}^2$};
	        
	        % Decoder
	        \node[rrnode_box]    (decoder_mean)  [above right = \h and 1.1*\w of aggregator] {$\bm{D_{\mu}}$};
	        \node[rrnode_box]    (decoder_L)   [below right = \h and 1.1*\w of aggregator] {$\bm{D_L}$};

	        \node[rrnode_data]  (w_mean) [above right = 0.2*\h and 0.7*\w of decoder_mean] {$\bm{\mu_{w_g}}$};
	        \node[rrnode_data]  (w_L)    [below right = 0.2*\h and 0.7*\w of decoder_L] {$\bm{L_{w_g}}$};

	        \node[rrnode_ag]   (reconstructor) [below right = \h and 1.3*\w of decoder_mean] {\acrshortpl{pdmp}};
	        
	        \node[rrnode_data]  (bc)            [above = 2*\h of reconstructor]{$\{t\},~\bm{y}_b,~\dot{\bm{y}}_b$};
	        
	        % Traj
	        \node[rrnode_data]  (y_mean)        [above right = 1.2*\h and 1.1*\w of reconstructor]{$\bm{\mu_{\Lambda}}$};
	        \node[rrnode_data]  (y_Sigma)       [below right = 1.3*\h and 1.1*\w of reconstructor]{$\bm{\Sigma_{\Lambda}}$};

	        \draw[-{Stealth[scale=0.6]}, ultra thick, color=\arrowblue] (ctx) -- (encoder_mean.west);
	        \draw[-{Stealth[scale=0.6]}, ultra thick, color=\arrowblue] (ctx) -- (encoder_var.west);

	        \draw[-{Stealth[scale=0.6]}, ultra thick, color=\arrowblue] (encoder_mean.east) -- (aggregator);
	        \draw[-{Stealth[scale=0.6]}, ultra thick, color=\arrowblue] (encoder_var.east) -- (aggregator);
	        
	        \draw[-{Stealth[scale=0.6]}, ultra thick, color=\arrowblue] (aggregator) -- (decoder_mean.west);
	        \draw[-{Stealth[scale=0.6]}, ultra thick, color=\arrowblue] (aggregator) -- (decoder_L.west);
	        
	        \draw[-{Stealth[scale=0.6]}, ultra thick,  color=\arrowblue] (decoder_mean.east) -- (reconstructor);
	        \draw[-{Stealth[scale=0.6]}, ultra thick,  color=\arrowblue] (decoder_L.east) -- (reconstructor);
	        
	        \draw[-{Stealth[scale=0.6]}, ultra thick, color=\arrowblue] (bc) -- (reconstructor);
	        
	        \draw[-{Stealth[scale=0.6]}, ultra thick, color=\arrowblue] (reconstructor) -- (y_mean);
	        \draw[-{Stealth[scale=0.6]}, ultra thick, color=\arrowblue] (reconstructor) -- (y_Sigma);
     	    \end{scope}
          \end{tikzpicture}
    }
    \vspace{-0.1cm}
\end{figure}
    
% \begin{figure}
%     \centering
%     \captionbox{Learn trajectory correlation through two randomly selected time points.\label{fig:cov_learning_2_points}}[0.28\textwidth]{\includegraphics[width=\linewidth]{figure/pdf/cov_learning/oc_dataset_small.pdf}}
% \end{figure}
\begin{figure}[t!]
\captionsetup[subfigure]{labelformat=empty}
\centering
% \hspace{-0.2cm}
\subcaptionbox{\label{fig:optimal_control}}
[.178\textwidth]{
    % \vspace{-0.4cm}
    \includegraphics[width=\linewidth]{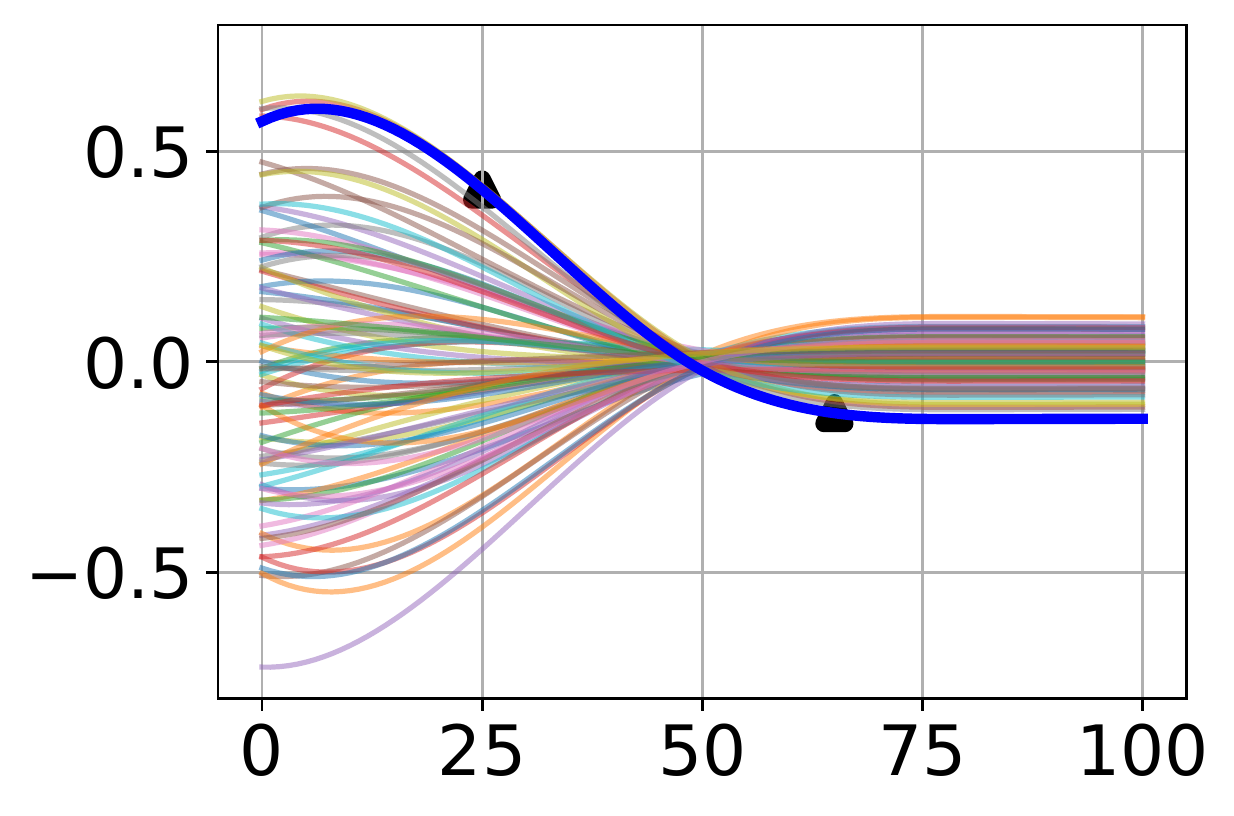}
}%
\hfill
\subcaptionbox{\label{fig:oc_promp}}
[.152\textwidth]{
    % \vspace{-0.4cm}
    \includegraphics[width=\linewidth]{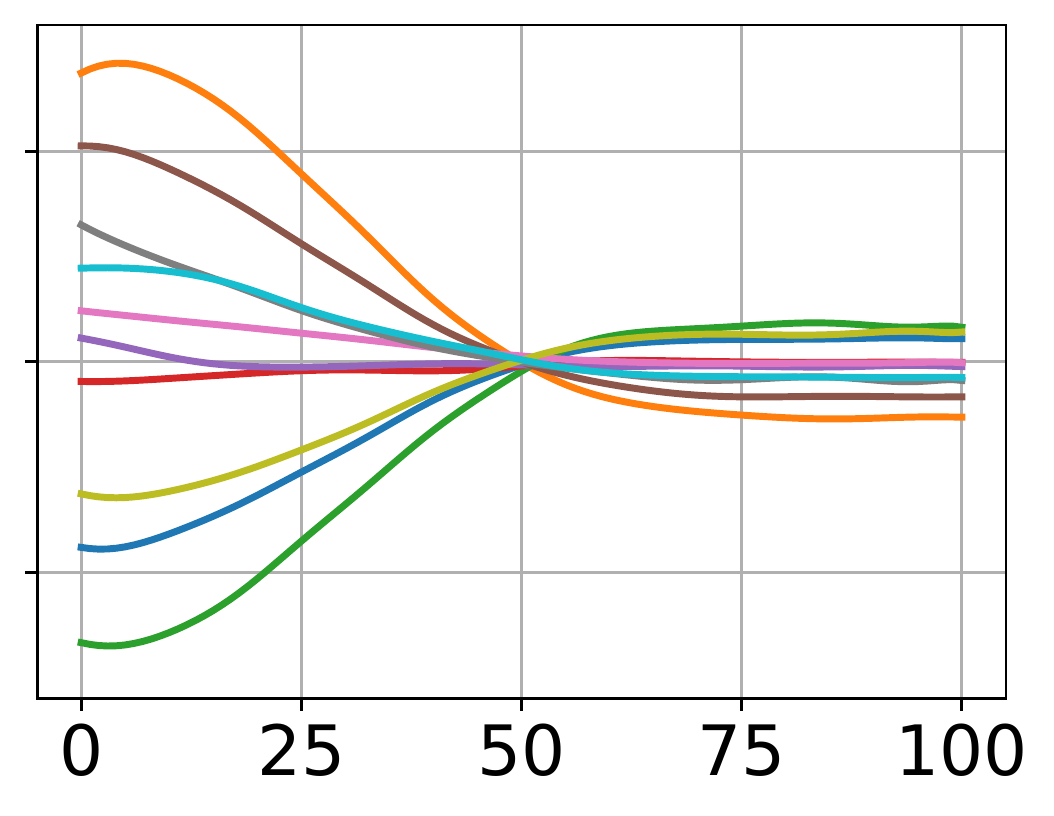}
}%
\hfill
\subcaptionbox{\label{fig:oc_promp_decouple}}
[.152\textwidth]{
    % \vspace{-0.4cm}
    \includegraphics[width=\linewidth]{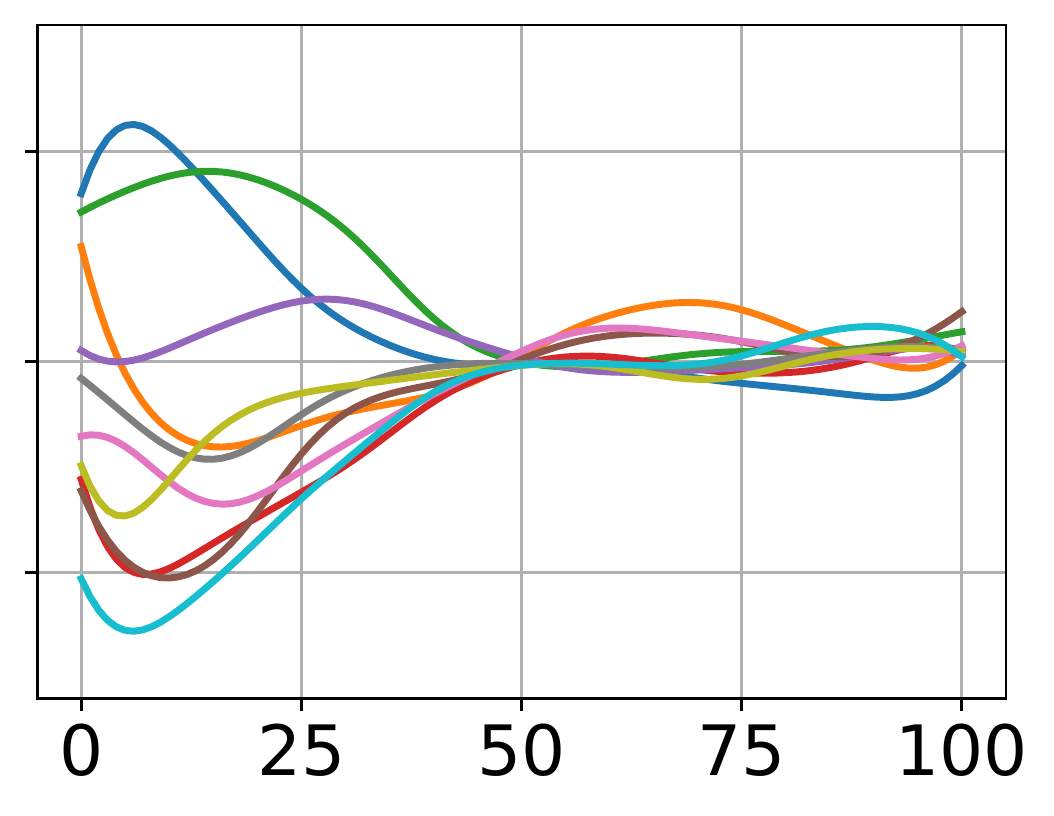}
}%
\vspace{-0.6cm}
\caption{Illustration of learning trajectory's temporal correlation using two time steps. {Left}: Trajectories in a dataset containing temporal correlation, which can be learned through two randomly selected time steps (triangle markers). {Middle}: Using two time steps jointly in Eq.(\ref{eq:rec_ll}) captures the temporal correlation correctly in the inference result. {Right}: Maximizing the likelihood of one single time step's value cannot learn the temporal correlation properly.\vspace{-0.0cm}}
\label{fig:temporal_cov}
\vspace{-0.2cm}
\end{figure}

\noindent
where $\bm{\mu}_{(t, t')_j}$ and $\bm{\Sigma}_{(t, t')_j}$ denote the mean and covariance of the joint distribution at the $j$-th paired time points which are obtained from the \acrshortpl{pdmp} decoder depicted in Figure \ref{fig:nmp_structure}. In Fig.~\ref{fig:temporal_cov}, we illustrate that our loss function can capture the temporal correlation. 
Further, an ablation study that only computes the likelihood of single-time steps results in a degenerated estimate of the temporal correlation. 

\section{Experiments}
\label{sec:experiments}
We highlight the advantages of our model and distinguish it from other methods through three experiments, which answer the following questions: 
\begin{enumerate*}[label=(\alph*)]
    \item Does our model produce high-quality trajectory distributions and samples using non-linear conditioning on high dimensional observations?
    \item Does it support online replanning and predicting a smooth trajectory distribution from the current robot state? 
    \item Can we condition on several partial observations and obtain a trajectory distribution that leverages the aggregated information?
\end{enumerate*} 
% \todo{In comparison to \acrshortpl{cnmp} \cite{seker2019conditional}, we show a clear advantage in the quality of the produced trajectory samples and the replanning capabilities.
% In comparison to, we show that our approach can generate a mean trajectory of similar quality but offers additional generative and statistical properties. }
We compare our method with state-of-the-art \acrshortpl{cnmp} \cite{seker2019conditional} and NN-based \acrshortpl{dmp} models \cite{gams2018deep, ridge2020training, bahl2020neural} on three digit-writing tasks using images as inputs, one simulated robot pushing task with complex physical interaction, and a real robot picking task with shifting object positions.
%%%%%%%%%%%%%%%%%%%%%%%%%%%%%%%%%%%%%%%%%%%%%%%%%%%%%%%%%%%%%%%%%%%%%%%%%%
% Digit writing
%%%%%%%%%%%%%%%%%%%%%%%%%%%%%%%%%%%%%%%%%%%%%%%%%%%%%%%%%%%%%%%%%%%%%%%%%%
\subsection{Learning Trajectory Distributions of the MNIST Digits}
\label{sbsec:exp_digit}
We use a synthetic-MNIST dataset \cite{ridge2020training} to showcase several properties of our method and advantages over the other approaches.
The dataset contains $20,000$ digit images in ten groups (0-9) and as prediction targets $3$ seconds trajectories with $2$ \acrshortpl{dof} each.
In our settings, we choose to use 25 basis functions per \acrshort{dof} and predict a 52-dim multivariate Gaussian distribution over \acrshortpl{pdmp} parameters (25 weights and 1 goal for each \acrshort{dof}). A 2-dim vector for the starting point of the trajectory is also predicted. 
We analyse our experiments based on three subtasks and show the spatial writing trajectories, i.\,e.\,, the x- and y-axis are spatial \acrshortpl{dof}.

\begin{figure}[t!]
\vspace{0.2cm}
\centering
\captionbox{The trajectories generated by different MP models. Only our model captures the temporal and DoF correlations.  \label{fig:digit_cov}}
[\linewidth]{
    \begin{minipage}[t!]{0.15\linewidth}
        \vspace{0.05cm}
        \includegraphics[width=\linewidth]{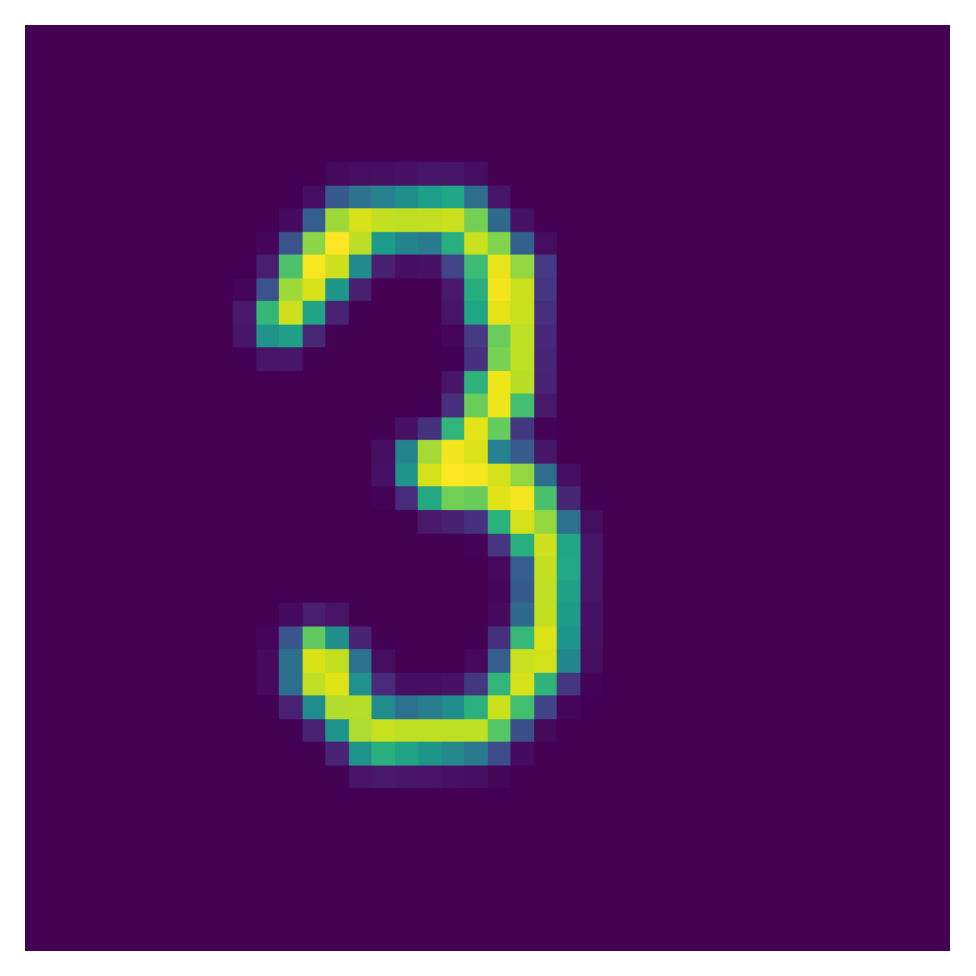}
        \label{fig:idmp_basis_pw}
        \vspace{-1.25mm}
        \caption*{\footnotesize Input}
        % \subcaption[]{Input}
    \end{minipage}
    \begin{minipage}[t!]{0.10\linewidth}
        \includegraphics[width=\linewidth]{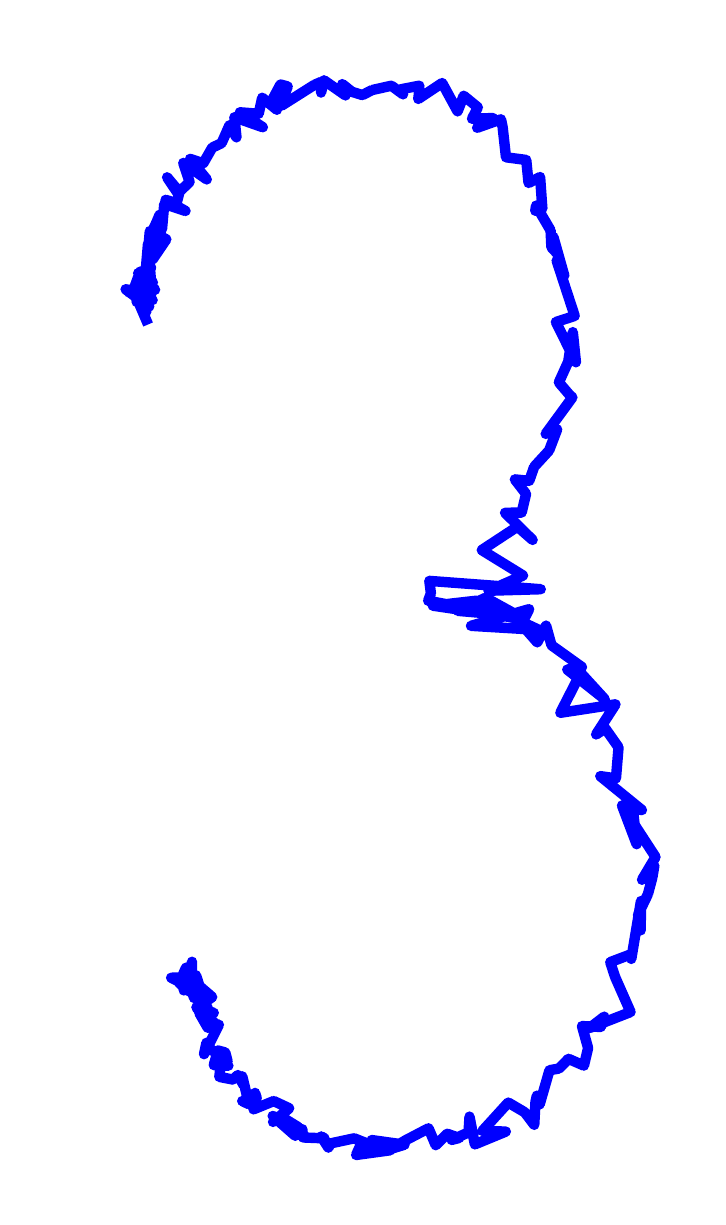}
        \label{fig:idmp_basis_pg}
        \vspace{-5.6mm}
        \caption*{\footnotesize \acrshort{cnmp} sample}
        % \subcaption[]{}
    \end{minipage}
    \begin{minipage}[t!]{0.10\linewidth}
        \includegraphics[width=\linewidth]{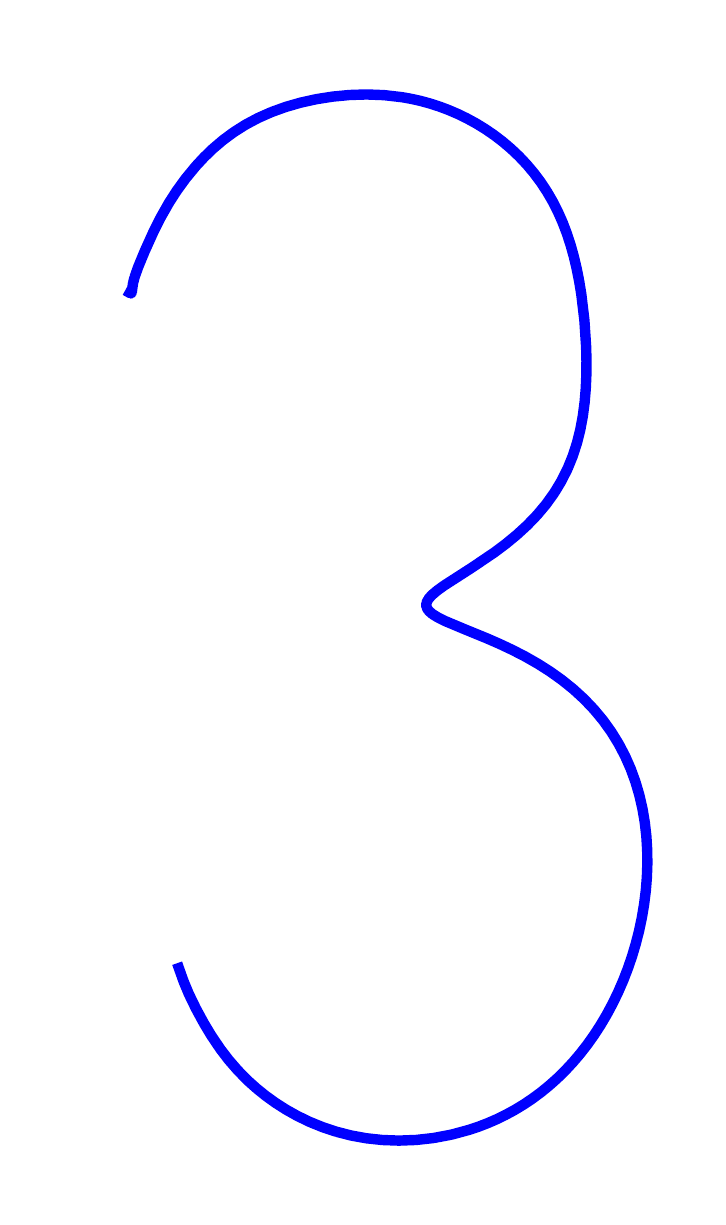}
        \label{fig:idmp_basis_vw}
        \vspace{-5.5mm}
        \caption*{\footnotesize $\phantom{...}$NN-$\phantom{..}$\acrshortpl{dmp}}
    \end{minipage}
    % \begin{minipage}[t!]{0.16\linewidth}
    %     \includegraphics[width=\linewidth]{figure/pdf/digit_one_image/nmp_std_3.pdf}
    %     \label{fig:idmp_basis_vg}
    %     \vspace{-4mm}
    %     % \caption*{Ours, std}
    % \end{minipage}
    \begin{minipage}[t!]{0.10\linewidth}
        \includegraphics[width=\linewidth]{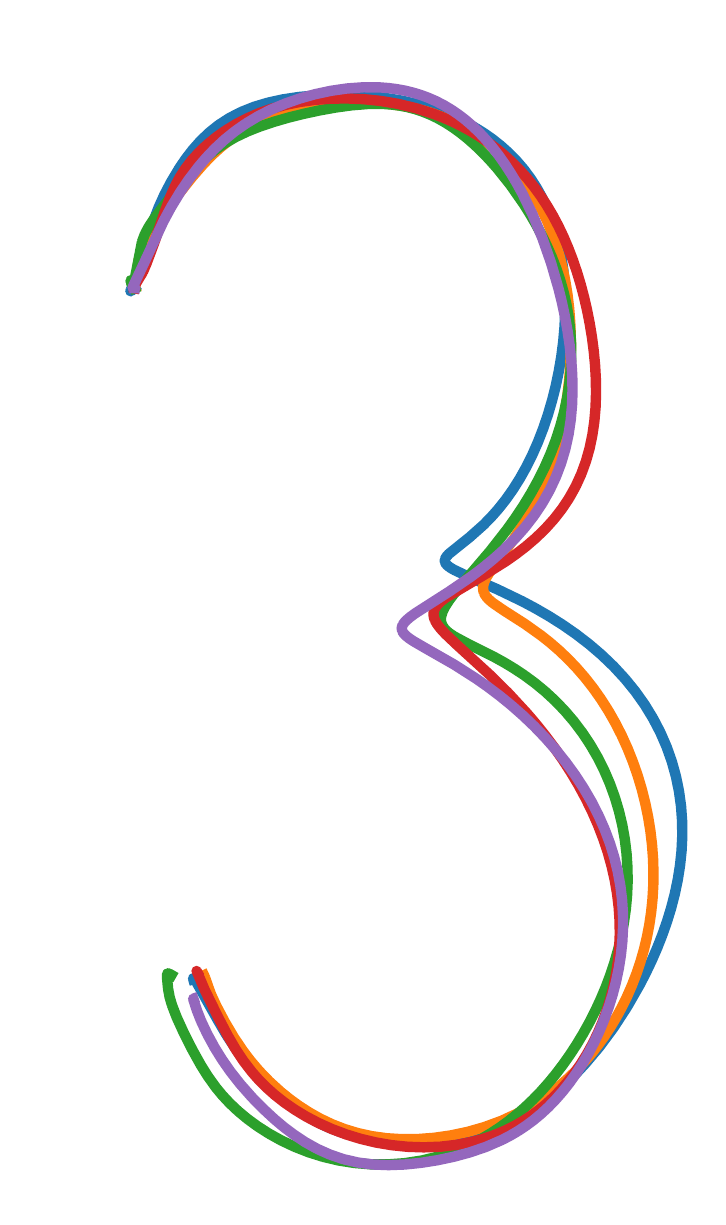}
        \label{fig:idmp_basis_vg}
        \vspace{-2.4mm}
        \caption*{\footnotesize Ours}
    \end{minipage}
    \begin{minipage}[t!]{0.15\linewidth}
        \vspace{0.05cm}
        \includegraphics[width=\linewidth]{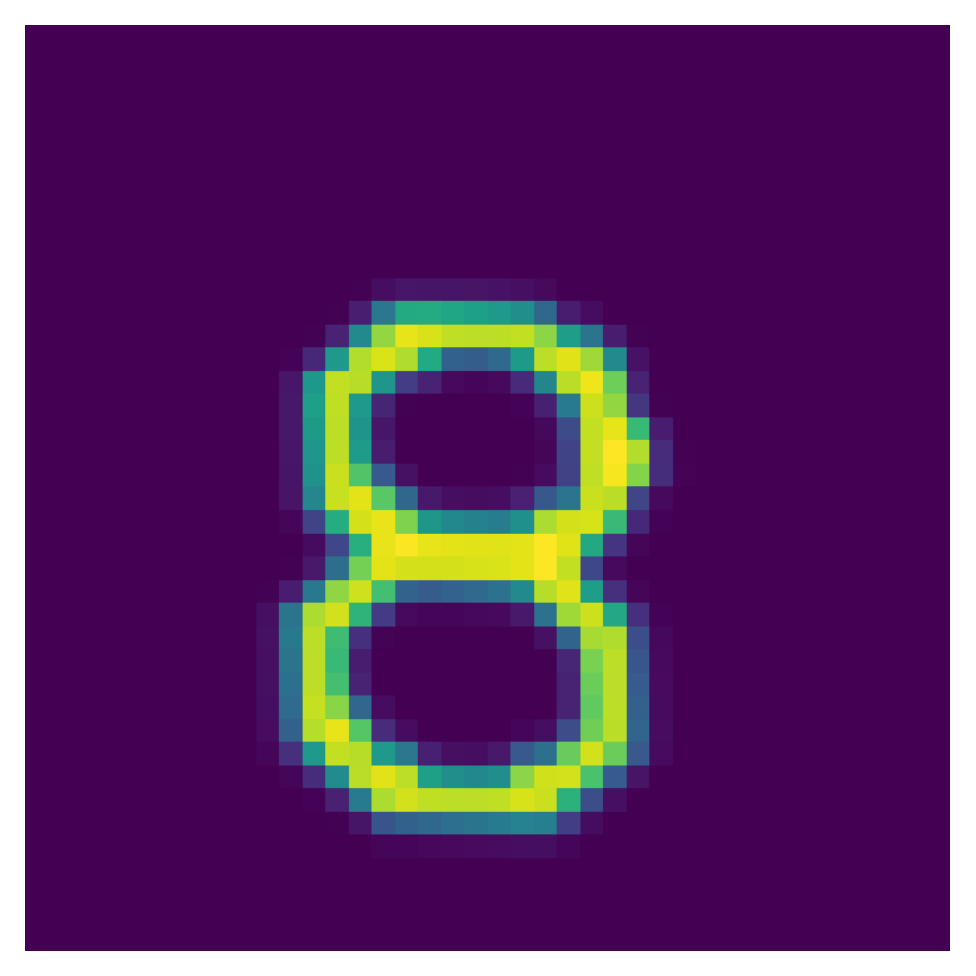}
        \label{fig:idmp_basis_pw}
        \vspace{-1.25mm}
        \caption*{\footnotesize Input}
    \end{minipage}
    \begin{minipage}[t!]{0.10\linewidth}
        \includegraphics[width=\linewidth]{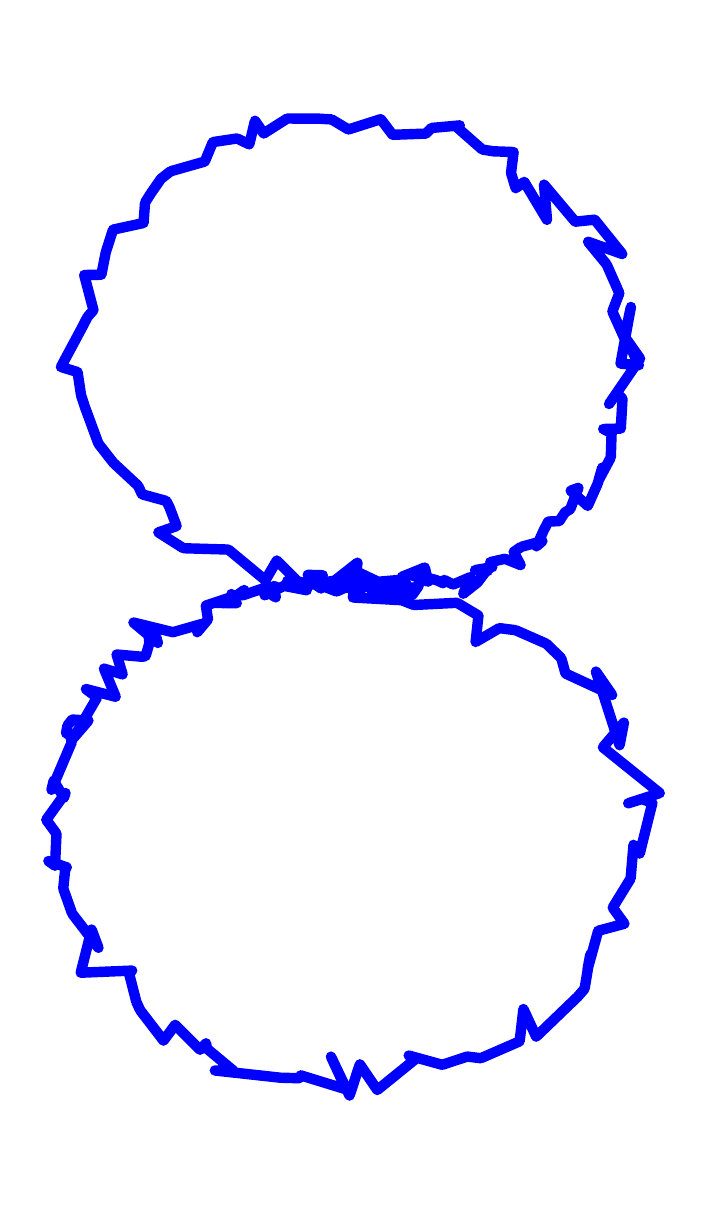}
        \label{fig:idmp_basis_pg}
        \vspace{-5.6mm}
        \caption*{\footnotesize \acrshort{cnmp} sample}
    \end{minipage}
    \begin{minipage}[t!]{0.10\linewidth}
        \includegraphics[width=\linewidth]{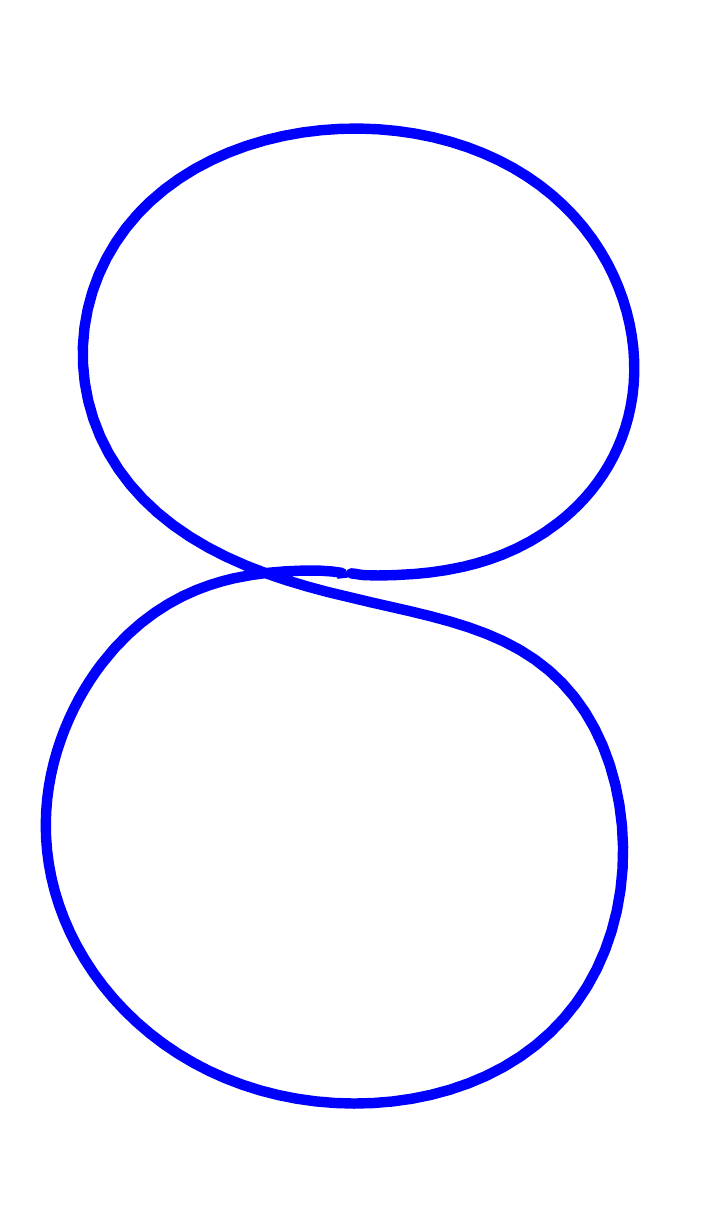}
        \label{fig:idmp_basis_vw}
        \vspace{-5.5mm}
        \caption*{\footnotesize $\phantom{...}$NN-$\phantom{..}$\acrshortpl{dmp}}
    \end{minipage}
    % \begin{minipage}[t!]{0.16\linewidth}
    %     \includegraphics[width=\linewidth]{figure/pdf/digit_one_image/nmp_std_8_new.pdf}
    %     \label{fig:idmp_basis_vg}
    %     \vspace{-3.6mm}
    %     \caption*{$\phantom{...}$Ours, $\phantom{....}$std}
    % \end{minipage}
    \begin{minipage}[t!]{0.10\linewidth}
        \includegraphics[width=\linewidth]{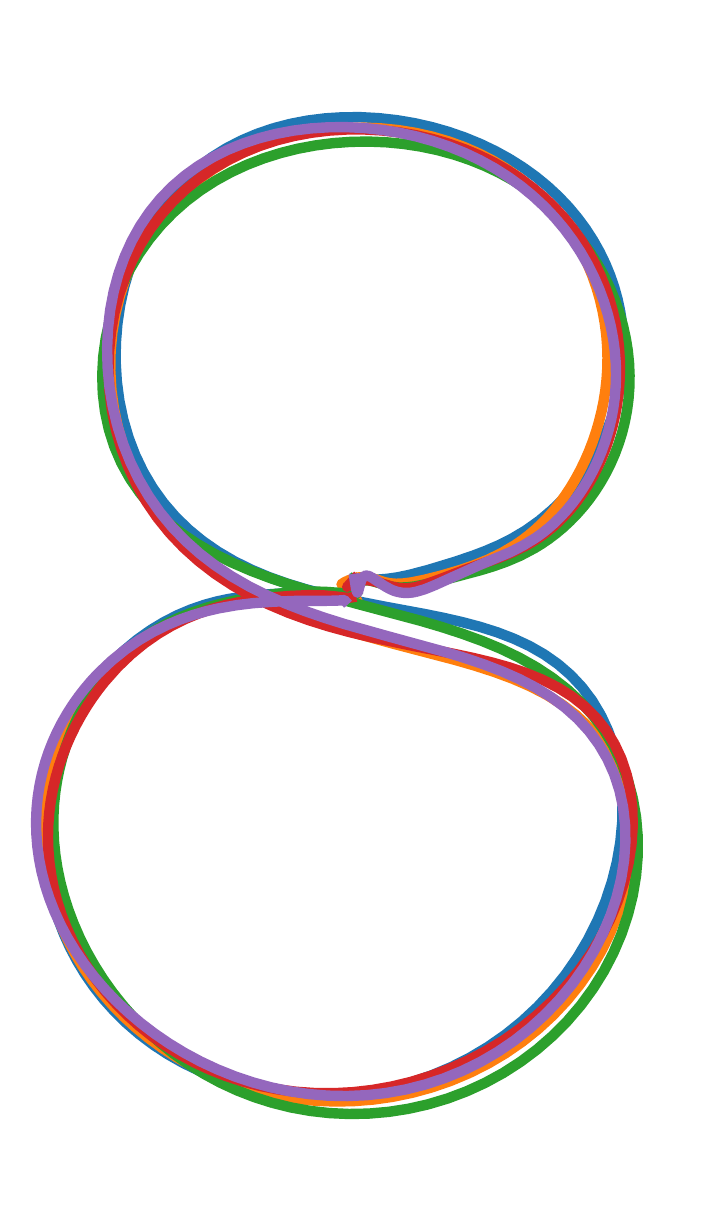}
        \label{fig:idmp_basis_vg}
        \vspace{-2.4mm}
        \caption*{\footnotesize Ours}
    \end{minipage}
}%
\end{figure}

\textbf{Comparison of Generative Models.}
As first task, we consider the prediction of trajectories using different methods given one image input. 
For \acrshortpl{cnmp} and \acrshortpl{pdmp}, we initially predict a distribution and then sample trajectories from it.
As shown in Fig.~\ref{fig:digit_cov}, \acrshortpl{cnmp} only model the mean and the isotropic variance per time step and thus fail to model correlations across time steps and dimensions. %Hence, trajectories sampled from \acrshortpl{cnmp}' predicted trajectory distribution are very noisy. We offer a discussion on variance learning in Appendix \ref{ap:cov}.
The NN-\acrshortpl{dmp} instead predicted a smooth trajectory, but do not learn any statistics over the trajectories and cannot generate samples. 
Our model, however, not only captures the correct shape of the trajectory but can also be used as a generative model to sample temporal and \acrshortpl{dof} consistent trajectories.  
To evaluate the speed of a full learning experiment, we replace the numerical integration of the NN-DMPs with our linear basis function model and keep the network architecture and \acrshort{mse} loss unchanged. The whole training time is reduced from 105 minutes to 10 minutes, which is approximately a speed-up by a factor of 10.

% \vspace{-0.2cm}
\begin{figure}[t!]
% \vspace{-5mm}
\centering
\captionbox{Given up to three noisy images, our \acrshortpl{pdmp} model aggregated the information and refined its prediction. \label{fig:digit_agg_one_img}}
[\linewidth]{
    \begin{minipage}[t!]{\linewidth}
        \includegraphics[width=\linewidth]{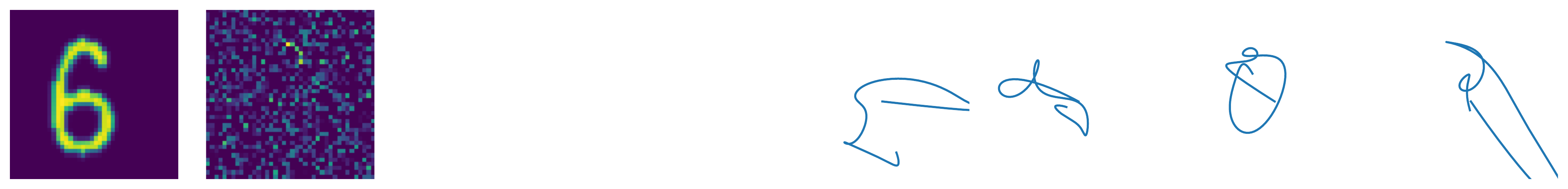}
        \vspace{-4mm}
        % \subcaption[]{}
    \end{minipage}
    \begin{minipage}[t!]{\linewidth}
        \includegraphics[width=\linewidth]{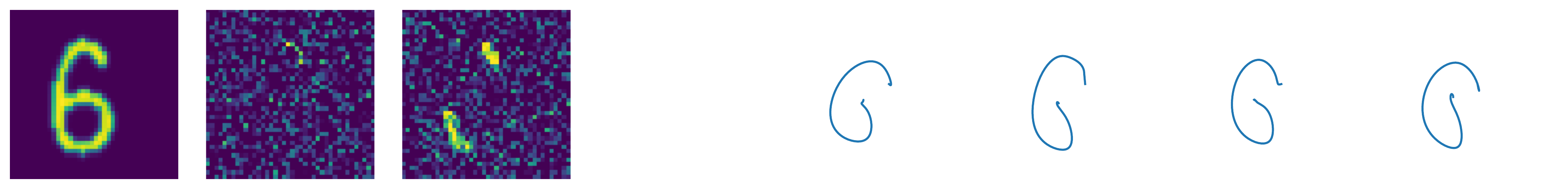}
        \vspace{-4mm}
        % \subcaption[]{}
    \end{minipage}
    \begin{minipage}[t!]{\linewidth}
        \includegraphics[width=\linewidth]{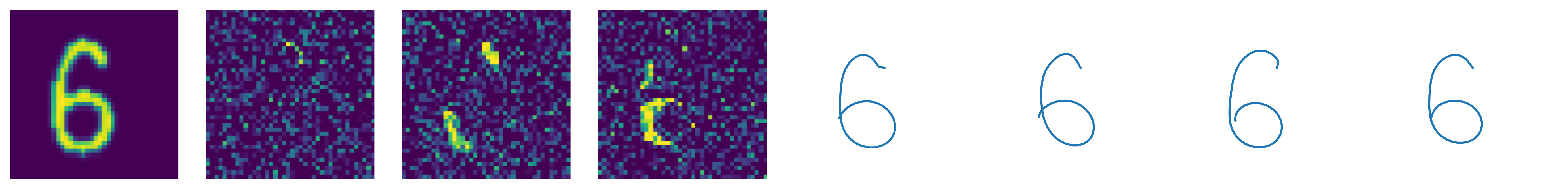}
        \vspace{-4mm}
        % \subcaption[]{}
    \end{minipage}
    \begin{minipage}[t!]{\linewidth}
        \includegraphics[width=\linewidth]{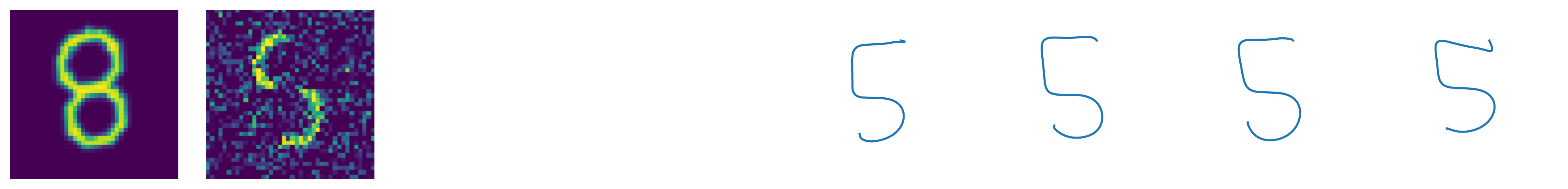}
        \vspace{-4mm}
        % \subcaption[]{}
    \end{minipage}
    \begin{minipage}[t!]{\linewidth}
        \includegraphics[width=\linewidth]{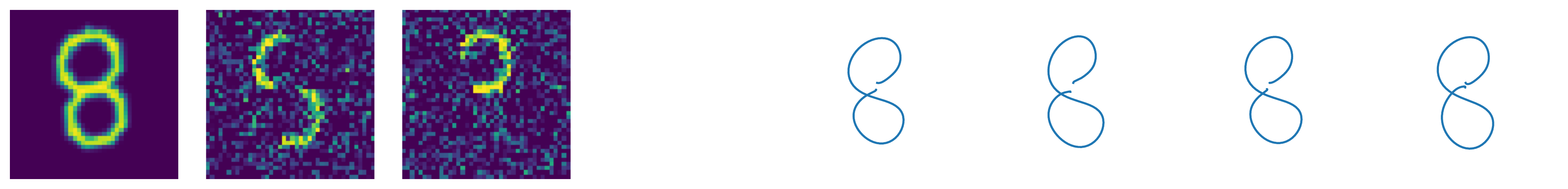}
        \vspace{-4mm}
        % \subcaption[]{}
    \end{minipage}
    \begin{minipage}[t!]{\linewidth}
        \includegraphics[width=\linewidth]{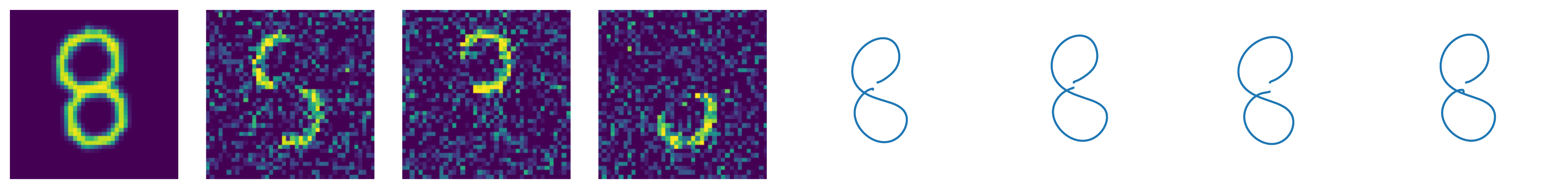}
        \vspace{-4mm}
        % \subcaption[]{}
    \end{minipage}
    \begin{minipage}[t!]{\linewidth}
        \begin{minipage}[t!]{0.12\linewidth}
            \centering
            \footnotesize{Origin}
        \end{minipage}
        \hfill
        \begin{minipage}[t!]{0.36\linewidth}
            \centering
            \footnotesize{Up to 3 noisy images}
        \end{minipage}
        \hfill
        \begin{minipage}[t!]{0.48\linewidth}
            \centering
            \footnotesize{\acrshortpl{pdmp} sampled trajectories}
        \end{minipage}
    \end{minipage}
}
\end{figure}
% \vspace{-0.2cm}
\begin{figure}[t!]
% \vspace{-5mm}
\centering
\captionbox{Write a digit in 3 steps. Our model received the first noisy image at the beginning while receiving the second and third at 25\% and 50\% of the execution time respectively. The trajectory gets replanned (blue $\rightarrow$ green $\rightarrow$ red) and executed accordingly.
% The dashed and solid lines distinguish trajectories before and after replanning, while their transitions are smooth. 
The dashed lines indicate the remaining, not executed trajectories.
\label{fig:digit_replan}}
[\linewidth]{
    \vspace{1mm}
    \begin{minipage}[t!]{\linewidth}
        \includegraphics[width=\linewidth]{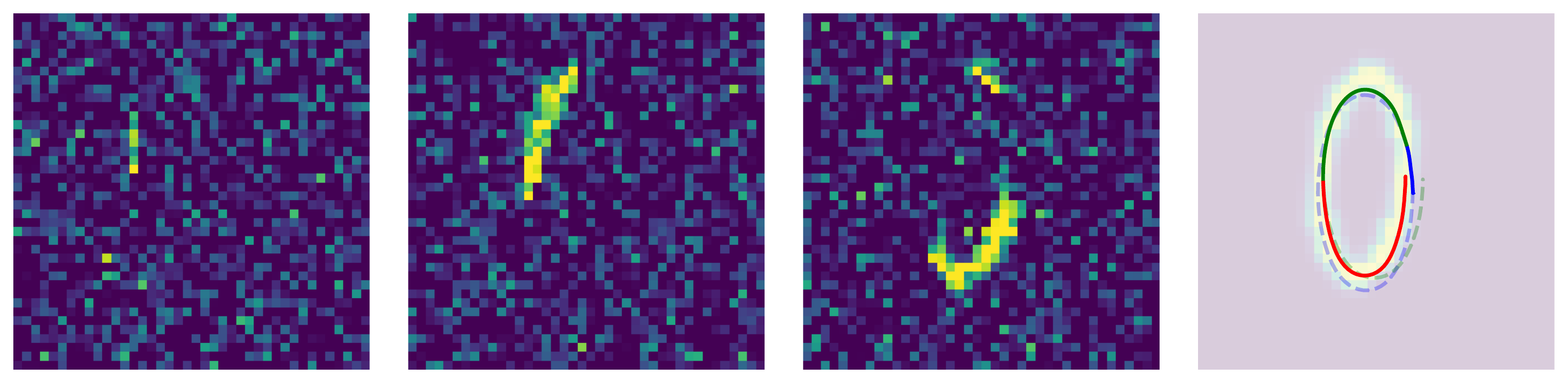}
        \vspace{-4mm}
        % \subcaption[]{}
    \end{minipage}
}
\vspace{-0.2cm}
\end{figure}

\textbf{Conditioning on Partial Observations.} 
For the second task, we show how our model leverages \acrlong{ba} to deal with multiple partial observations.
We repeatedly add synthetic noise to each original image such that we obtain three noisy images for each digit. The random noise masks occlude most of the image, leaving only a few original pixels visible. Our model successfully aggregated the information contained in the different observations and provided a trajectory distribution refinement over the predicted digit. Fig.~\ref{fig:digit_agg_one_img} shows that the trajectories appear random without sufficient information about the digit, but become refined once more information is added.

\textbf{Online Replanning.}
Finally, we show how our method benefits from its dynamical system in online replanning once new information is provided. To this end, we provide new observations to the method during the execution of the trajectory.
The same noisy images as in the previous experiment of the target digit are provided at 0\%, 25\%, and 50\% of the execution time.
For illustration purposes, we trained the models only on images containing one type of digit. 
As shown in Fig.~\ref{fig:digit_replan}, each additional observation increases the accuracy of the trajectory, while the new trajectories transition smoothly at the replanning points.

\subsection{Pushing and Replanning with Complex Contact}
\label{sbsec:robot_push}

\begin{figure}[t!]
\vspace{0.2cm}
    \centering
    \captionsetup[subfigure]{labelformat=empty}
    \subcaptionbox{}[0.32\linewidth]{
        \includegraphics[width=\linewidth]{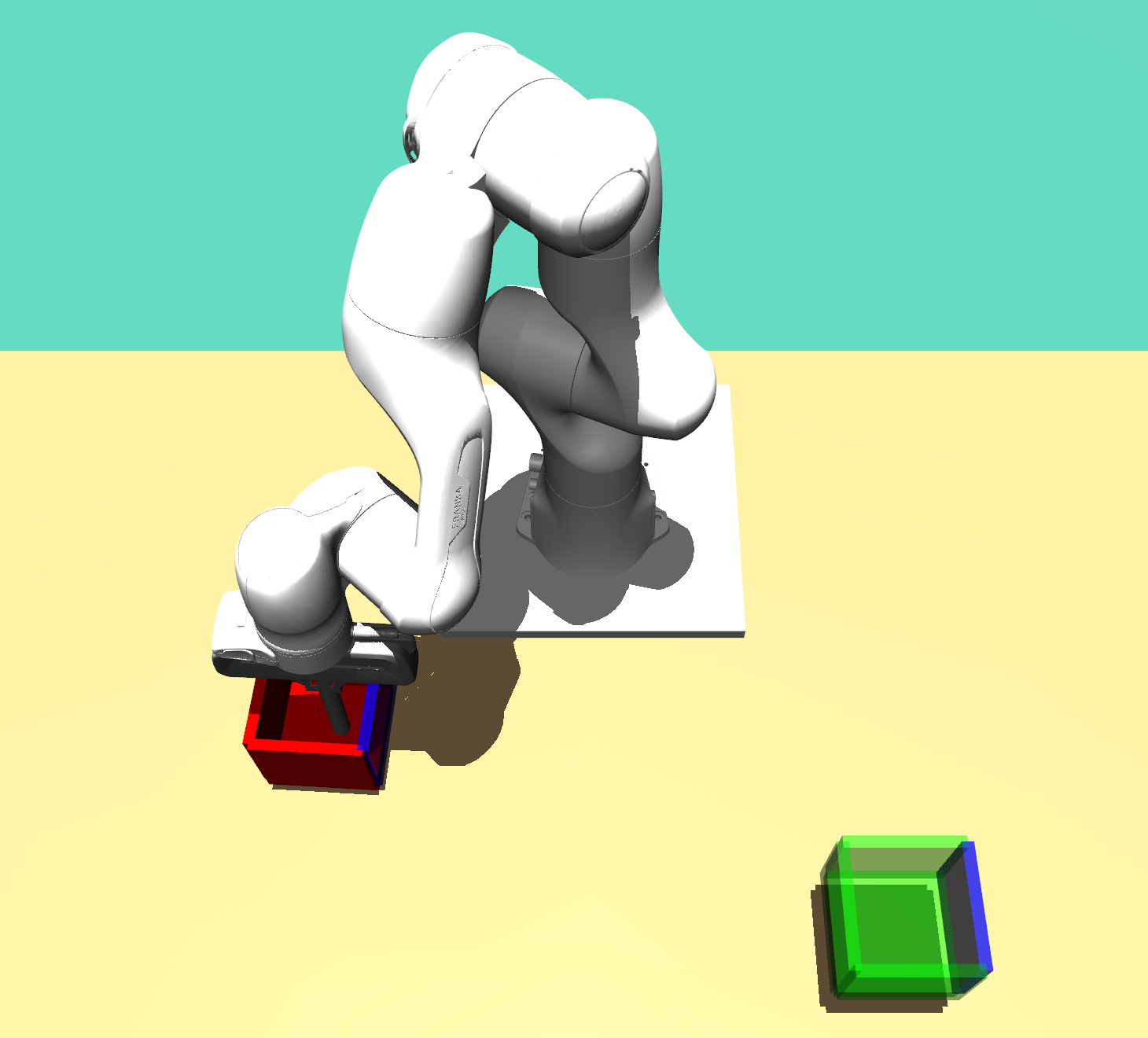}
    }
    \subcaptionbox{}[0.32\linewidth]{
        \includegraphics[width=\linewidth]{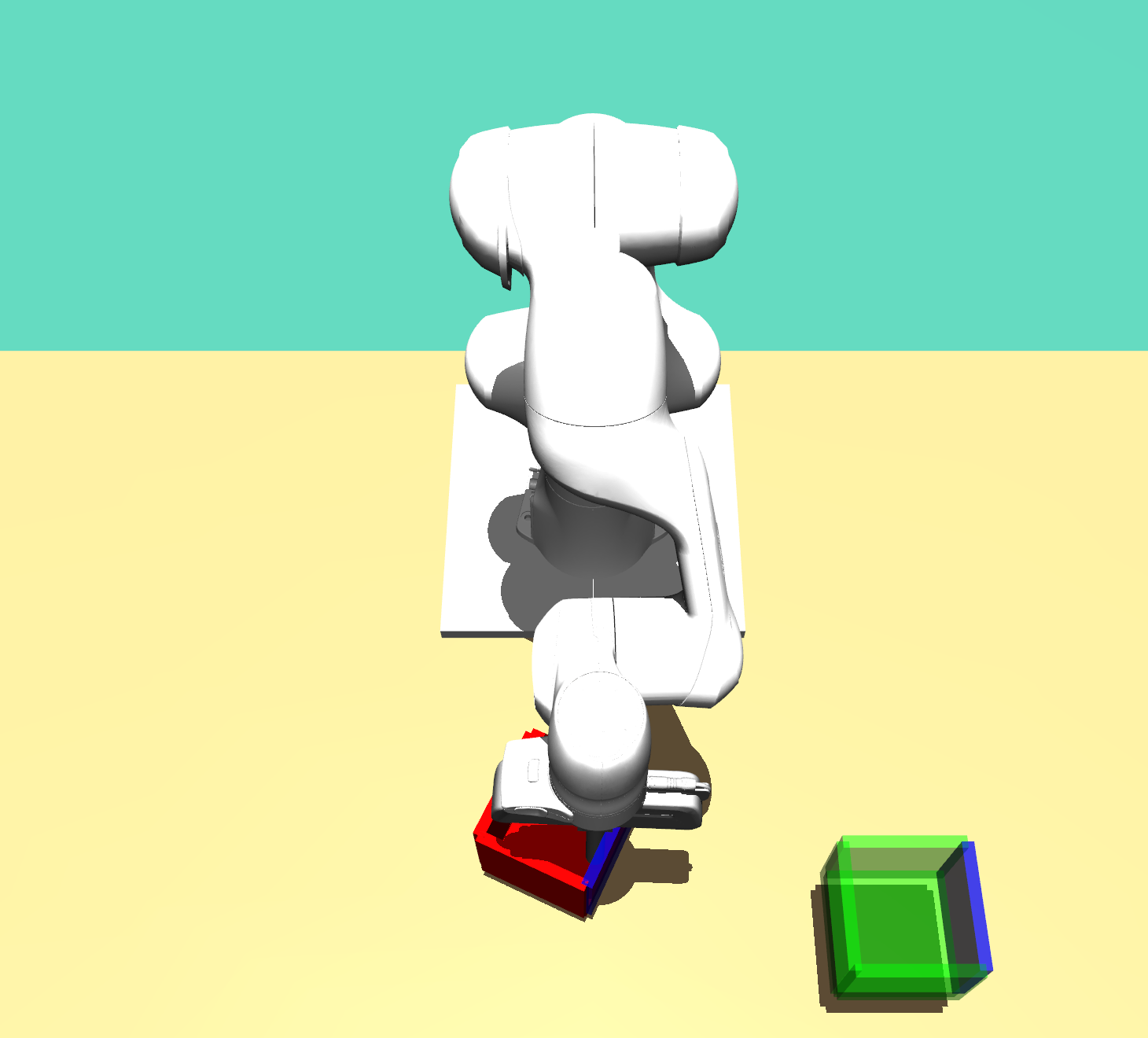}
    }
    \subcaptionbox{}[0.32\linewidth]{
        \includegraphics[width=\linewidth]{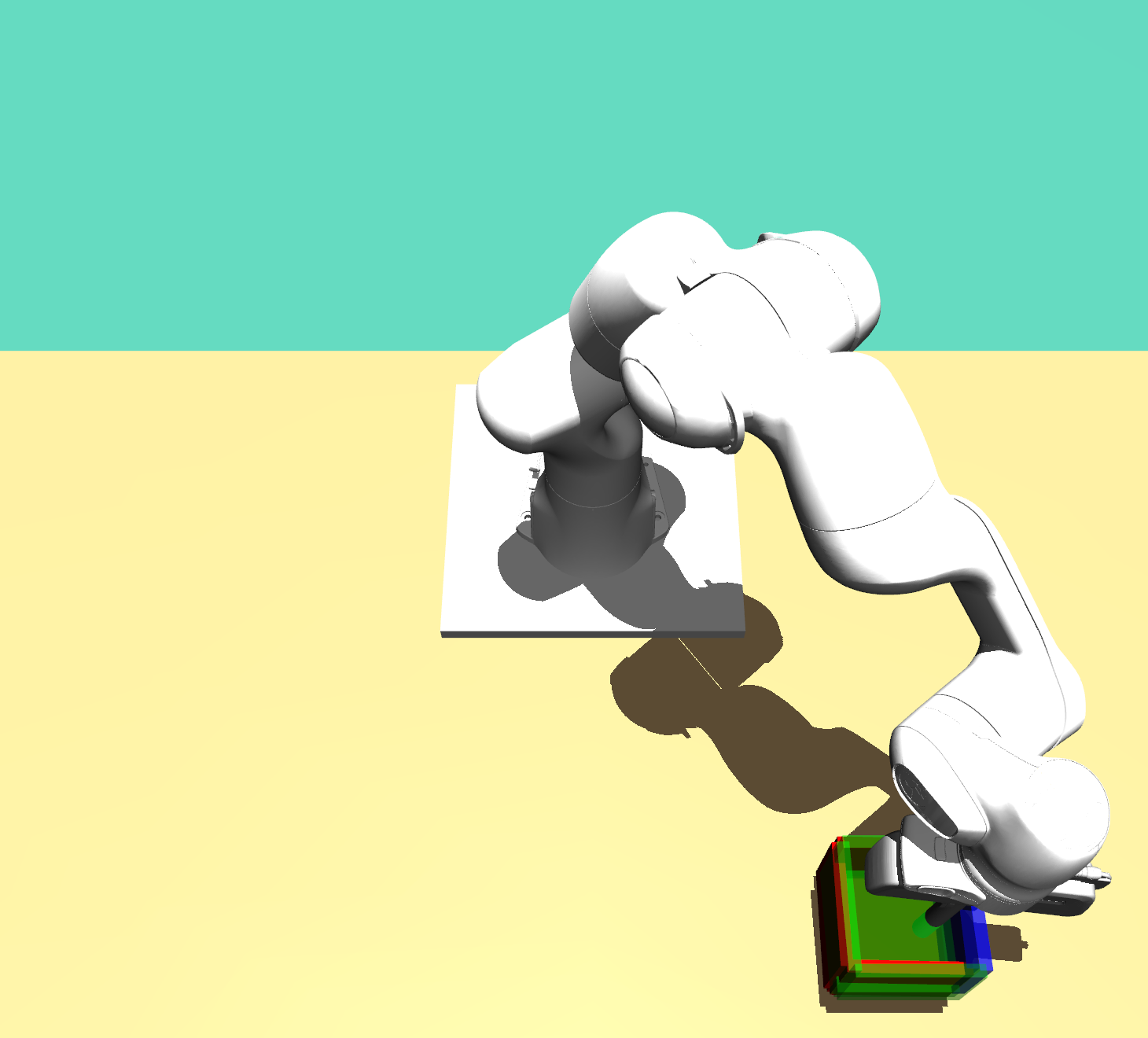}
    }
    \vspace{-0.4cm}
    \caption{Pushing an empty box (red) from a randomized initial state to a given target position and orientation (green) in Mujoco \cite{todorov2012mujoco}. }
    \label{fig:robot_pushing_screenshot}
\end{figure}

We conduct a simulated robot experiment to prove that our approach can deal with rich physical interactions and online conditioning on environmental variables.
We use a Franka Panda robot in Mujoco \cite{todorov2012mujoco} to push a box with its end-effector (cf.\ Fig.~\ref{fig:robot_pushing_screenshot}).
The box has a square shape and is empty. The end-effector is equipped with a peg and starts in the box. 
Using a simple teleoperation interface, we control the end-effector's $x$ and $y$ position with a mouse, which is then translated into the joint angles using inverse kinematics (IK).
We collected 225 demos, each $3$ seconds long, moving the box from a random initial state to a desired target location and orientation.
The box is hard to control, especially the orientation, due to the non-linear interaction of its walls and corners with the peg. 
The learning procedure is as follows.
At each iteration, we randomly choose a replanning time point and use the recorded box position, orientation, as well as the actual position and velocity of the robot end-effector in the past $0.1$s as observations. 
We predict a desired end-effector trajectory distribution in the next $0.5$s, using \acrshortpl{pdmp} with 25 basis functions per \acrshort{dof}, and maximize the log-likelihood of the demonstrated trajectory. In the inference phase, our model directly interacts with the simulator, pushing a box from an unseen initial state.
We executed either the mean or the sampled trajectory of the predicted trajectory distribution. 
We compare our method with 
\begin{enumerate*}[label=(\alph*)]
    \item \acrshortpl{cnmp} only using the mean,
    \item \acrshortpl{cnmp} with sampling,
    \item ablated \acrshortpl{pdmp} without replanning, and
    \item a vanilla behavior cloning network mapping observations to actions at each time step.
\end{enumerate*}

% \noindent
We evaluate the error between the final and target states of the box, as well as the average squared acceleration (ASA, the smaller the better) as a trajectory smoothness metric. The result is shown in Table~\ref{tab:robot_push}.
Our model can reproduce the box-pushing task with similar accuracy as the demos and achieves excellent levels of smoothness. 
The performance of the \acrshortpl{cnmp}' mean prediction yields a similar pushing result to our model. 
However, the trajectory samples are, as expected, unable to solve the task. 
In addition, \acrshortpl{cnmp} exhibit noticeable jumps and significant accelerations at the transition points of the trajectory. 
The sampled trajectories are lacking temporal and \acrshortpl{dof} correlation, leading to an extremely low level of smoothness.
The ablated case of our model cannot solve the task properly, indicating that replanning is crucial to adjust the movement in such a contact-rich scenario. 
The comparably high smoothness is achieved by the lack of replanning in the trajectory.
The state-action-based behavioral cloning using an \acrshort{mse} loss cannot capture the temporal correlation between the box and the robot's movement, and thus performs poorly.
It is worth mentioning that using NN-based \acrshortpl{dmp} in the current task is equivalent to training our model with the \acrshort{mse} loss, which will lead to the same result but a faster speed.
% However, our model will be trained faster, as the only difference is the way to generate the trajectory, i.\,e.\, through numerical integration vs. through linear basis functions. 

\renewcommand{\arraystretch}{1.5}
\setcounter{table}{1}
\setlength{\aboverulesep}{0pt}
\setlength{\belowrulesep}{0pt}
\captionsetup{font=small}
\begin{figure}[t]
    \begin{minipage}[b]{\linewidth}
        \scriptsize
        \centering
        \vspace{1.2mm}
        \captionof{table}{Evaluation of different settings using (a) Mean and std of the absolute error (x-/y-position and orientation along z-axis) of the final box state to its target state, as well as (b) the average squared acceleration (ASA) as smoothness metric. Each model is trained 20 times using different random seeds. \label{tab:robot_push}}
        % \vspace{-0.2cm}
        \begin{tabular}{cccccc}
            \toprule
            & \textbf{Replan} &\textbf{x (mm)} & \textbf{y (mm)} & \textbf{z-ori. (deg)} & \textbf{ASA}\\
            % \hline
            \textbf{Demos} & - & \centering 4.2 $\pm$ 2.9 & \centering 4.8 $\pm$ 3.4 & \hfil 1.76 $\pm$ 1.28 & - \\
            \hline
            \textbf{\acrshort{pdmp}} & mean & \centering \textbf{9.2 $\pm$ 8.2}& \centering \textbf{7.1 $\pm$ 7.2} & \hfil \textbf{4.3 $\pm$ 5.7} & 3.3\\
            % \textbf{\acrshort{pdmp} (std)} & mean & \centering 16.7 $\pm$ 15.0& \centering 14.5 $\pm$ 19.2 & \hfil 10.0 $\pm$ 16.2 \\
            \textbf{\acrshort{cnmp}} & mean & \centering 9.5 $\pm$ 9.2& \centering 7.5 $\pm$ 9.7 & \hfil 4.7 $\pm$ 9.0 & 800.5\\
            \hline
            \textbf{\acrshort{pdmp}} & sample &\centering 19.0 $\pm$ 31.3&\centering 17.6 $\pm$ 20.1&\hfil 12.1 $\pm$ 21.1 & 13.1\\
            \textbf{\acrshort{cnmp}} & sample & \centering 72.4 $\pm$ 13.7& \centering 80.8 $\pm$ 21.5 & \hfil 20.8 $\pm$ 14.4 & 2.7e4\\
            \hline
            \textbf{\acrshort{pdmp}} & $\times$ & \centering 62.0 $\pm$ 53.9& \centering 55.9 $\pm$ 38.8 & \hfil 33.6 $\pm$ 28.4 & \textbf{0.3}\\
            \textbf{B. C.} &per step& \centering 35.8 $\pm$ 34.0& \centering 94.9 $\pm$ 14.2 & \hfil 21.2 $\pm$ 24.0 & 7.1\\
            \bottomrule
        \end{tabular}
        % \vspace{0.3cm}
    \end{minipage}
%  \vspace{-0.4cm}
\end{figure}

\begin{figure}[t!]
    \begin{center}
    \subcaptionbox{\acrshortpl{pdmp}' mean\label{fig:prodmp_ymean}}[0.522\linewidth]{
        \includegraphics[width=\linewidth]{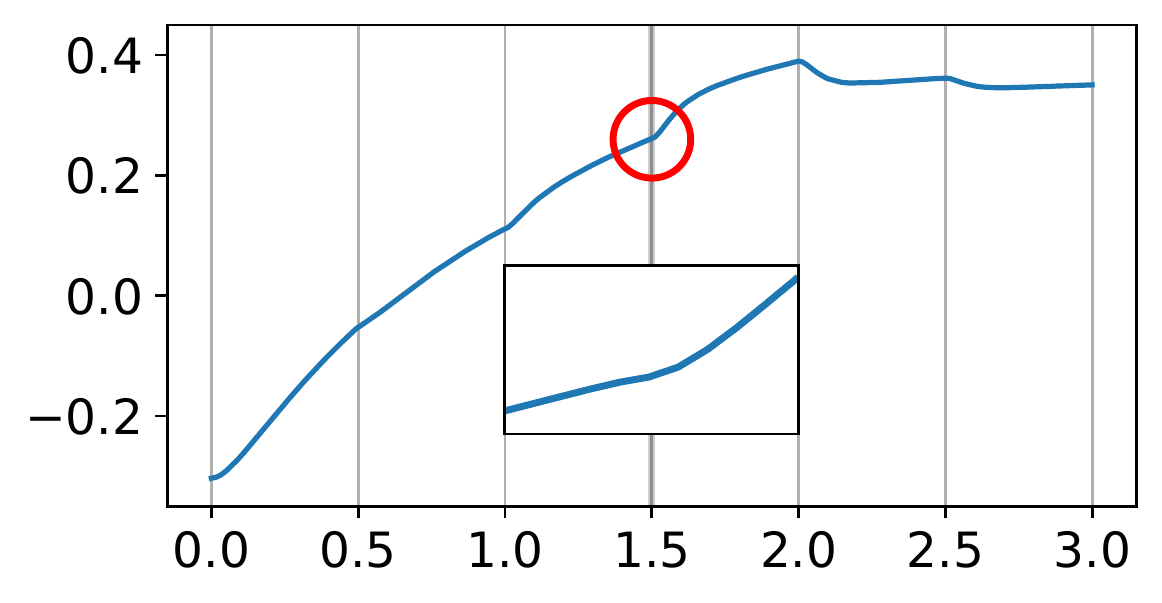}}
    \hfill
    \subcaptionbox{\acrshortpl{cnmp}' mean\label{fig:cnmp_ymean}}[0.465\linewidth]{
        \includegraphics[width=\linewidth]{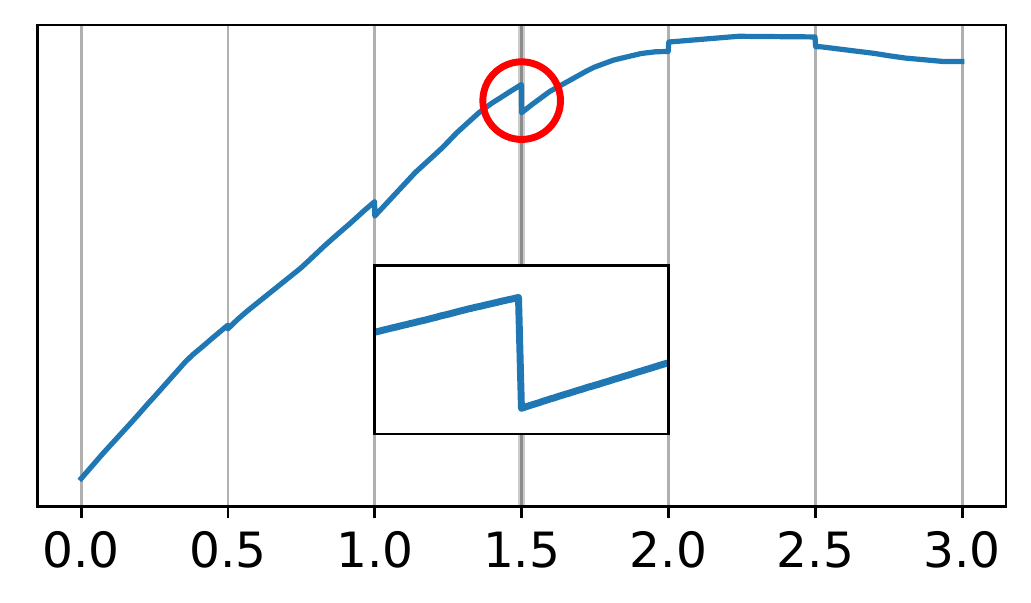}}
  \end{center}
  \vspace{-0.25cm}
  \caption{Replanning trajectories comparison of robot pushing. x-axis: time [s], y-axis: end-effector y position [m]. 
  Each trajectory is a concatenation of 6 replanning segments, 0.5 second long each. Our model is smooth at the replanning time steps (vertical lines), while the \acrshortpl{cnmp} are discontinuous when replanning.}
  \label{fig:cnmp_push}
  \vspace{-0.2cm}
\end{figure}

\subsection{Object Picking with Dynamic Positional Shift}
\begin{figure}[t!]
\centering
\vspace{0.1cm}
\subcaptionbox{Real Setup\label{fig:robot_pick_setup}}
[.232\textwidth]{
    \includegraphics[width=\linewidth]{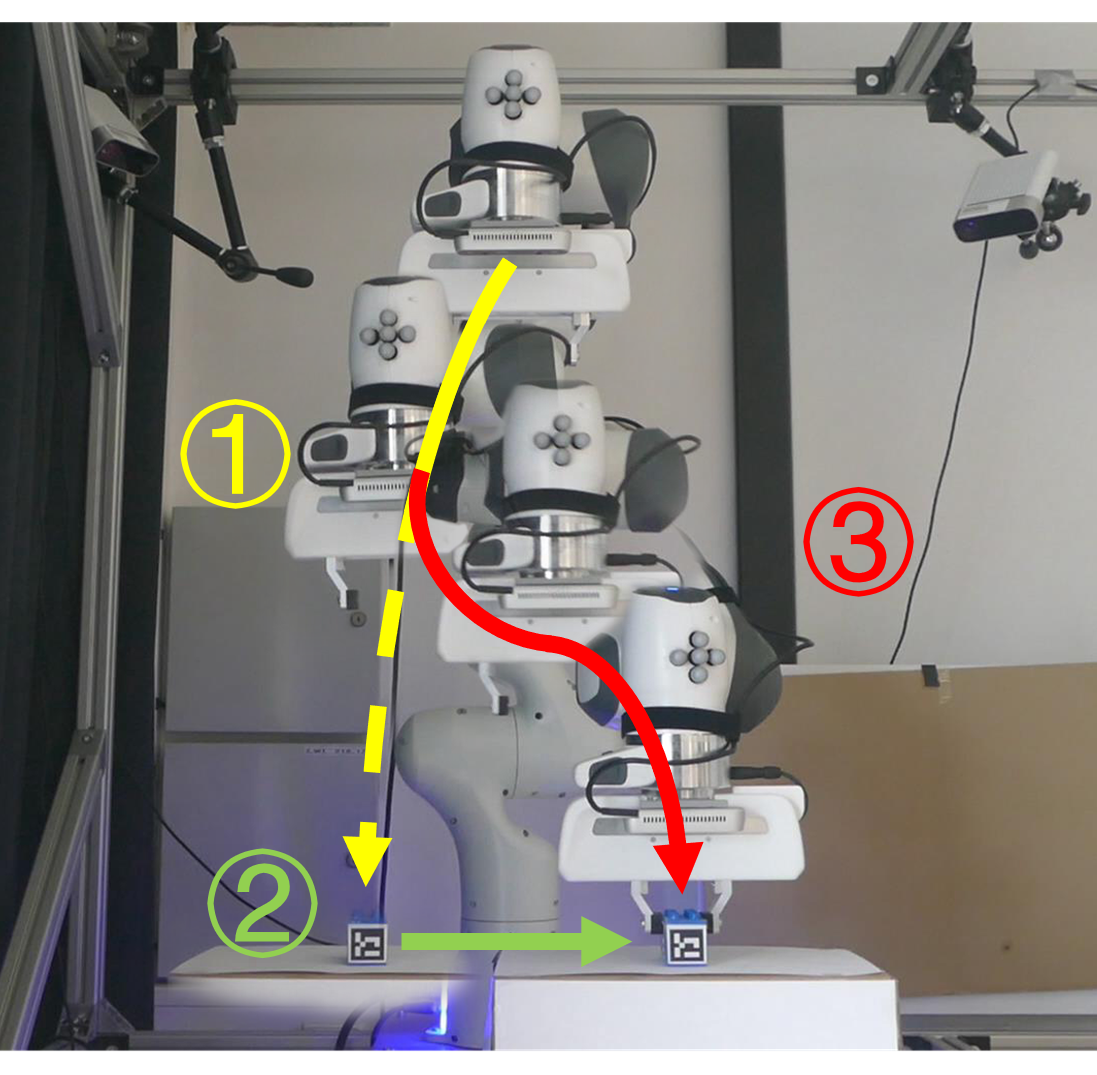}
}%
\hfill
\subcaptionbox{The 3rd joint position\label{fig:robot_pick_replanning}}
[.21\textwidth]{
    \includegraphics[width=0.18\textwidth]{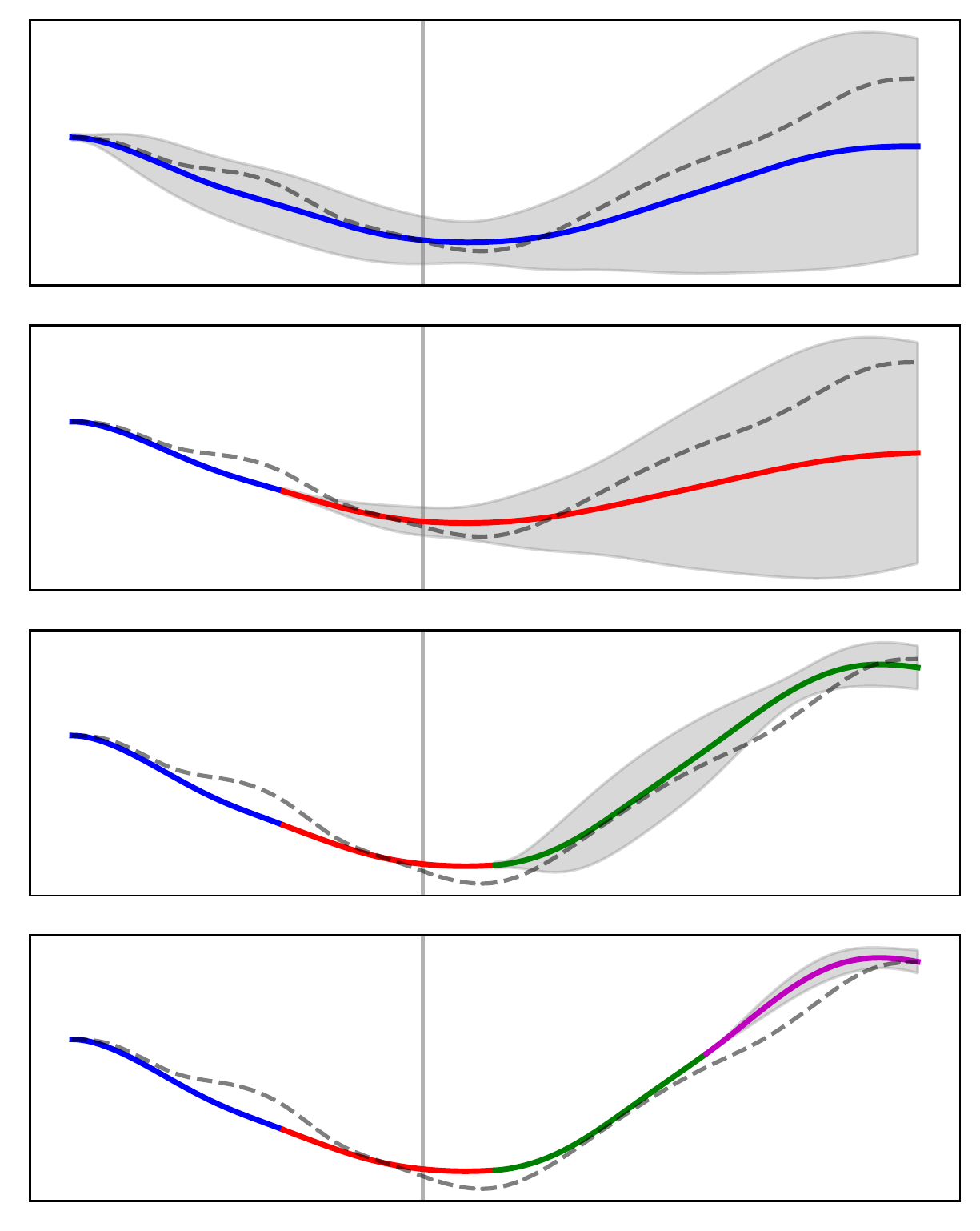}
}%
\caption{
(a) The robot is picking an object initialized on the left. During the movement, the object was shifted to the right, and the robot replanned a new trajectory from its current state to the new object position without interruption. 
(b) For each episode, every 1 second our model predicts a new distribution using its current state and the latest box position. Before the shift (vertical line), the predictions contain high variance. After the shift, the variance decreases intensely. The dashed line denotes the human's demo.
}
\label{fig:robot_pick}
\end{figure}

Finally, we learn an object-picking task in a 7-\acrshortpl{dof} joint space. 
A Franka Panda robot picks an object in its workspace (cf.\ Fig.~\ref{fig:robot_pick_setup}).
During execution, the object shifts to a new position, and the robot must plan a new trajectory from its current position to pick up the object without interruption.
We collected 100 human demos, each 4 seconds long, using the same teleoperation interface as in Section~\ref{sbsec:robot_push}. In each demo, the human moves the robot end-effector to the object and adjusts the movement when the object shifts. The demonstrated joint trajectory is computed using IK. 
% We show one of the joints' in Fig. \ref{fig:robot_pick_demo}.
After each second, the actual object position and the robot state are given to \acrshortpl{pdmp} to replan a trajectory distribution starting from the current robot state. We evaluated the picking success rate of our model using the
\begin{enumerate*}[label=(\alph*)]
    \item mean prediction, and
    \item trajectory samples in a simulated and a real setup
\end{enumerate*}. 
The box position in the real setup is captured by a camera system using ArUco ROS \cite{salinas2015aruco}.
% Please do not change the hard coded table index!
Our model learns picking and replanning movement directly from the data and achieves a high success rate in simulation, as shown in TABLE~\ref{tab:robot_pick}. Due to the camera delay in the real setup, the performance decreases slightly.
% Please do not change the hard coded table index!
Fig.~\ref{fig:robot_pick_replanning} shows four predicted trajectory distributions in one sequence. Our model predicts the trajectory distribution with a high variance at the beginning and decreases the predicted variance once it observes the shift of the object. 

\begin{table}[t]
    \scriptsize
    \centering
    \caption{Success rate and shift distance of 100 picks}
    \label{tab:robot_pick}
    % \vspace{-0.3cm}
    \begin{tabular}{ccccc}
        \toprule
        100 picks of & Sim+mean & Sim+sample & Real+mean & Real+sample\\
        \hline
        \textbf{Success rate} & 99\% & 82\% & 87\% & 75\% \\
        \textbf{Shift distance} & \multicolumn{2}{c}{22.8 $\pm$5.6} & 21.7 $\pm$9.7 & 21.0 $\pm$8.0\\
        \bottomrule
    \end{tabular}
\end{table}
\section{Conclusion}
\label{sec:conclusion}
We presented \acrshortpl{pdmp}, a unified framework fusing dynamic and probabilistic movement primitives. \acrshortpl{pdmp} recovered a linear basis-function representation for the trajectories by solving the \acrshort{ode} of the dynamical system. 
This way, we can easily represent trajectory distributions that adhere to boundary conditions defined by the current robot position and velocity as well as generate smooth trajectories when replanning. 
Further, we built a neural aggregation model for non-linear iterative conditioning and found a solution to learn full trajectory covariances with fewer resources. 
Our deep embedded \acrshortpl{pdmp} achieved smoothness, goal convergence, trajectory correlation modeling, non-linear trajectory conditioning, and online replanning in one model.
% We conclude and compare the properties of our and other approaches in Table~\ref{tab:model_comparison}. 
For future work, we will extend our approach to reinforcement learning, also considering force profiles that need to be applied. 
% \textbf{Limitations.}
% Our deep embedding assumes a single mode over the task's prior. Thus, given no observation, the prediction cannot distinguish different data modes or generate meaningful samples, e.g. the first row in Fig. \ref{fig:digit_agg_one_img}. We believe using, e.g., Gaussian Mixture Models or Variational Auto-Encoders can solve this issue. Besides, learning full covariance needs a relatively larger network and a longer time in training. The loss function, in this case, is noisier than learning the std-only case or using the MSE loss. We had to add gradient-clipping to avoid the training collapse. For future work, we will deploy \acrshortpl{pdmp} in RL, as our observing-replanning mechanism can naturally fit into the RL's framework, using accumulated observations and avoid replanning at each single time-step.

\section{Acknowledgement}
The authors acknowledge support by the state of Baden-Württemberg through bwHPC.
%%%%%%%%%%%%%%%%%%%%%%%%%%%%%%%%%%%%%%%%%%%%%%%%%%%%%%%%%%%%%%%%%%%%%%%%%%%%%%%%

\renewcommand{\bibfont}{\footnotesize} % for IEEE bibfont size
\printbibliography

%%%%%%%%%%%%%%%%%%%%%%%%%%%%%%%%%%%%%%%%%%%%%%%%%%%%%%%%%%%%%%%%%%%%%%%%%%%%%%%%
\newpage
\appendices
{\balance
\def\arrowblue{black!40!blue}
\def\arrowgreen{black!40!green}
\usetikzlibrary{arrows,matrix,positioning,patterns,decorations.pathreplacing,calc}
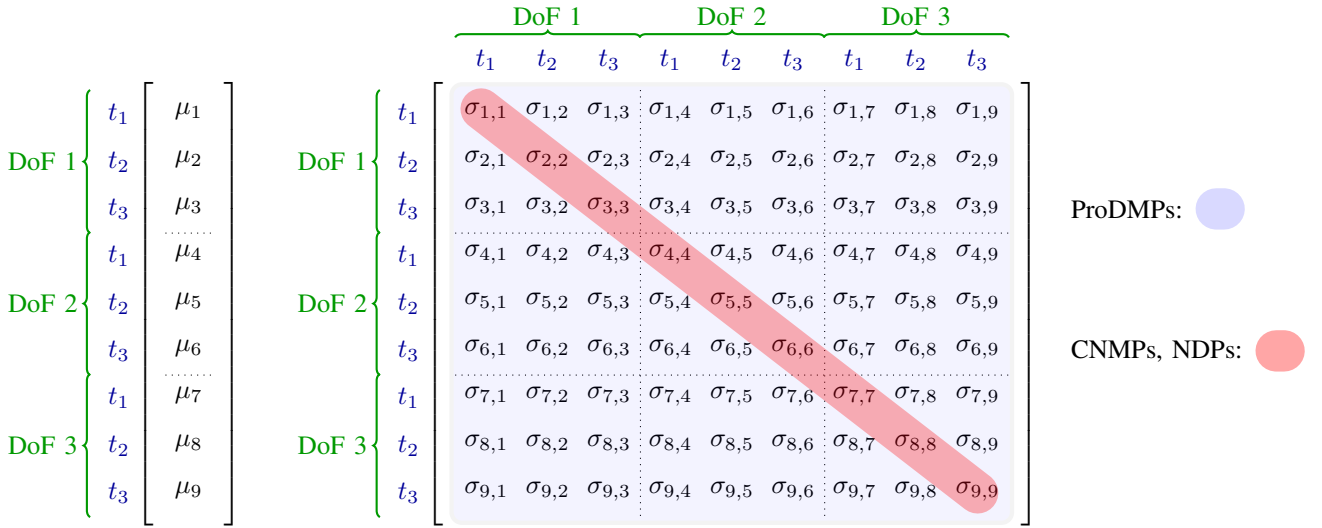
\begin{figure*}[b!]
\vspace{2cm}
    \centering
    \begin{tikzpicture}[baseline,decoration=brace, rrnode_nn/.style={rectangle, draw, minimum size=5mm, rounded corners=1mm, scale=\sca},]
        \node[ ]  (start) at (0,0) {$ $};
        \matrix [matrix of math nodes, left delimiter=[, right delimiter={]},text depth=1.25ex,
        text height=1.25ex,] (mean) [right=2cm of start]
        {
            \mu_{1}\\ 
            \mu_{2}\\ 
            \mu_{3}\\ 
            \mu_{4}\\ 
            \mu_{5}\\ 
            \mu_{6}\\ 
            \mu_{7}\\ 
            \mu_{8}\\ 
            \mu_{9}\\ 
        };
        \draw [dotted] (mean-3-1.south west) -- (mean-3-1.south east);
        \draw [dotted] (mean-6-1.south west) -- (mean-6-1.south east);
        \draw[decorate,transform canvas={xshift=-10mm},thick, color=\arrowgreen] (mean-3-1.south west) -- node[left=1pt] {DoF 1} (mean-1-1.north west); % right
        \draw[decorate,transform canvas={xshift=-10mm},thick, color=\arrowgreen] (mean-6-1.south west) -- node[left=1pt] {DoF 2} (mean-4-1.north west); % right
        \draw[decorate,transform canvas={xshift=-10mm},thick, color=\arrowgreen] (mean-9-1.south west) -- node[left=1pt] {DoF 3} (mean-7-1.north west); % right
        
        % time
        \node [left=1em of mean-1-1.west, color=\arrowblue] {$t_1$};
        \node [left=1em of mean-2-1.west, color=\arrowblue] {$t_2$};
        \node [left=1em of mean-3-1.west, color=\arrowblue] {$t_3$};
        \node [left=1em of mean-4-1.west, color=\arrowblue] {$t_1$};
        \node [left=1em of mean-5-1.west, color=\arrowblue] {$t_2$};
        \node [left=1em of mean-6-1.west, color=\arrowblue] {$t_3$};
        \node [left=1em of mean-7-1.west, color=\arrowblue] {$t_1$};
        \node [left=1em of mean-8-1.west, color=\arrowblue] {$t_2$};
        \node [left=1em of mean-9-1.west, color=\arrowblue] {$t_3$};
        
        \matrix [matrix of math nodes, left delimiter=[, right delimiter={]}, text depth=1.25ex,
        text height=1.25ex,] (m) [right=3cm of mean]
        {
            \sigma_{1,1} & \sigma_{1,2} & \sigma_{1,3} & \sigma_{1,4} & \sigma_{1,5} & \sigma_{1,6}& \sigma_{1,7}& \sigma_{1,8}& \sigma_{1,9} \\ 
            \sigma_{2,1} & \sigma_{2,2} & \sigma_{2,3} & \sigma_{2,4} & \sigma_{2,5} & \sigma_{2,6}& \sigma_{2,7}& \sigma_{2,8}& \sigma_{2,9} \\ 
            \sigma_{3,1} & \sigma_{3,2} & \sigma_{3,3} & \sigma_{3,4} & \sigma_{3,5} & \sigma_{3,6}& \sigma_{3,7}& \sigma_{3,8}& \sigma_{3,9} \\ 
            \sigma_{4,1} & \sigma_{4,2} & \sigma_{4,3} & \sigma_{4,4} & \sigma_{4,5} & \sigma_{4,6}& \sigma_{4,7}& \sigma_{4,8}& \sigma_{4,9} \\ 
            \sigma_{5,1} & \sigma_{5,2} & \sigma_{5,3} & \sigma_{5,4} & \sigma_{5,5} & \sigma_{5,6}& \sigma_{5,7}& \sigma_{5,8}& \sigma_{5,9} \\ 
            \sigma_{6,1} & \sigma_{6,2} & \sigma_{6,3} & \sigma_{6,4} & \sigma_{6,5} & \sigma_{6,6}& \sigma_{6,7}& \sigma_{6,8}& \sigma_{6,9} \\ 
            \sigma_{7,1} & \sigma_{7,2} & \sigma_{7,3} & \sigma_{7,4} & \sigma_{7,5} & \sigma_{7,6}& \sigma_{7,7}& \sigma_{7,8}& \sigma_{7,9} \\ 
            \sigma_{8,1} & \sigma_{8,2} & \sigma_{8,3} & \sigma_{8,4} & \sigma_{8,5} & \sigma_{8,6}& \sigma_{8,7}& \sigma_{8,8}& \sigma_{8,9} \\ 
            \sigma_{9,1} & \sigma_{9,2} & \sigma_{9,3} & \sigma_{9,4} & \sigma_{9,5} & \sigma_{9,6}& \sigma_{9,7}& \sigma_{9,8}& \sigma_{9,9} \\ 
        };
        %simple rectangle
        % \draw (m-3-1.south west) rectangle (m-1-3.north east);
        %fancy blue rectangle
                  
        \draw [dotted] (m-3-1.south west) -- (m-3-9.south east);
        \draw [dotted] (m-6-1.south west) -- (m-6-9.south east);
        \draw [dotted] (m-1-3.north east) -- (m-9-3.south east);
        \draw [dotted] (m-1-6.north east) -- (m-9-6.south east);

        \draw[decorate,transform canvas={yshift=7mm},thick, color=\arrowgreen] (m-1-1.north west) -- node[above=1pt] {DoF 1} (m-1-3.north east); % right
        \draw[decorate,transform canvas={yshift=7mm},thick, color=\arrowgreen] (m-1-4.north west) -- node[above=1pt] {DoF 2} (m-1-6.north east); % right
        \draw[decorate,transform canvas={yshift=7mm},thick, color=\arrowgreen] (m-1-7.north west) -- node[above=1pt] {DoF 3} (m-1-9.north east); % right
        \draw[decorate,transform canvas={xshift=-10mm},thick, color=\arrowgreen] (m-3-1.south west) -- node[left=1pt] {DoF 1} (m-1-1.north west); % right
        \draw[decorate,transform canvas={xshift=-10mm},thick, color=\arrowgreen] (m-6-1.south west) -- node[left=1pt] {DoF 2} (m-4-1.north west); % right
        \draw[decorate,transform canvas={xshift=-10mm},thick, color=\arrowgreen] (m-9-1.south west) -- node[left=1pt] {DoF 3} (m-7-1.north west); % right
        
        % times
        \node [above=0.4em of m-1-1.north, color=\arrowblue] {$t_1$};
        \node [above=0.4em of m-1-2.north, color=\arrowblue] {$t_2$};
        \node [above=0.4em of m-1-3.north, color=\arrowblue] {$t_3$};
        \node [above=0.4em of m-1-4.north, color=\arrowblue] {$t_1$};
        \node [above=0.4em of m-1-5.north, color=\arrowblue] {$t_2$};
        \node [above=0.4em of m-1-6.north, color=\arrowblue] {$t_3$};
        \node [above=0.4em of m-1-7.north, color=\arrowblue] {$t_1$};
        \node [above=0.4em of m-1-8.north, color=\arrowblue] {$t_2$};
        \node [above=0.4em of m-1-9.north, color=\arrowblue] {$t_3$};
        
        \node [left=1em of m-1-1.west, color=\arrowblue] {$t_1$};
        \node [left=1em of m-2-1.west, color=\arrowblue] {$t_2$};
        \node [left=1em of m-3-1.west, color=\arrowblue] {$t_3$};
        \node [left=1em of m-4-1.west, color=\arrowblue] {$t_1$};
        \node [left=1em of m-5-1.west, color=\arrowblue] {$t_2$};
        \node [left=1em of m-6-1.west, color=\arrowblue] {$t_3$};
        \node [left=1em of m-7-1.west, color=\arrowblue] {$t_1$};
        \node [left=1em of m-8-1.west, color=\arrowblue] {$t_2$};
        \node [left=1em of m-9-1.west, color=\arrowblue] {$t_3$};        
        
        % STD
        % \draw[rounded corners,ultra thick, draw=black, fill=green, opacity=0.1] (m-1-1.north west) rectangle (m-3-3.south east);  
        % \draw[rounded corners,ultra thick, draw=black, fill=green, opacity=0.1] (m-4-4.north west) rectangle (m-6-6.south east);  
        % \draw[rounded corners,ultra thick, draw=black, fill=green, opacity=0.1] (m-7-7.north west) rectangle (m-9-9.south east);  
        
        % COV
        \draw[rounded corners,ultra thick, draw=black, fill=blue, opacity=0.05] (17.2em, -8.3em) rectangle (38.4em, 8.3em);  
        
        %CNMP 
        \draw[opacity=.3,line width=5.5mm,line cap=round, color=red] (m-1-1.center)+(-0.2em, 0.2em) -- (m-9-9.center)+(0.2em, -3em);
        
        %Legend
        \node (cnmp) [right=2em of m-6-9.east] {\acrshortpl{cnmp}, \acrshortpl{ndp}: };
        \matrix [matrix of math nodes, text depth=1.25ex, text height=1.25ex,] (cnmpl) [right = 1mm of cnmp]{\phantom{o}\\};
        \draw[opacity=.35,line width=5.5mm,line cap=round, color=red] (cnmpl-1-1.center)+(-0.3em, 0.0em) -- (cnmpl-1-1.center)+(0.5em, 0.0em);
        
        \node (pdmps) [right=2em of m-3-9.east] {\acrshortpl{pdmp}:};
        \matrix [matrix of math nodes, text depth=1.25ex, text height=1.25ex,] (pdmpsl) [right = 1mm of pdmps]{\phantom{o}\\};
        \draw[opacity=.15,line width=5.5mm,line cap=round, color=blue] (pdmpsl-1-1.center)+(-0.3em, 0.0em) -- (pdmpsl-1-1.center)+(0.5em, 0.0em);
        
        % \node (pdmpc) [below=2em of m-9-7.west] {\acrshortpl{pdmp} (ablated, std):};
        % \matrix [matrix of math nodes, text depth=1.25ex, text height=1.25ex,] (pdmpcl) [right = 1mm of pdmpc]{\phantom{o}\\};
        % \draw[opacity=.2,line width=5.5mm,line cap=round, color=green,  opacity=0.3] (pdmpcl-1-1.center)+(-0.3em, 0.0em) -- (pdmpcl-1-1.center)+(0.5em, 0.0em);

    \end{tikzpicture}    
    \vspace{-0.5cm}
    \caption{Element-wise representation of a multi-variant Gaussian distributed trajectory. This example trajectory has coupled 3 time steps and 3 \acrshortpl{dof}. Left: the trajectory mean vector. Right: the trajectory covariance matrix. We add time steps labels ({\color{\arrowblue}$ t_1, t_2, t_3$}) and \acrshortpl{dof} labels ({\color{\arrowgreen}DoF1, DoF2, DoF3}) to indicate each element's meaning, e.g. $\mu_5$ is the trajectory mean value of the 2nd \acrshort{dof} at the 2nd time step, and $\sigma_{1,9}$ represents the covariance between the trajectory value of the 1st \acrshort{dof} at the 1st time step and the trajectory value of the 3rd \acrshort{dof} at the 3rd time step. Our method learns the entire trajectory covariance matrix (elements in light blue), while the other method \cite{bahl2020neural, seker2019conditional} learns only the isotropic variance (elements red).}
    \label{fig:traj_full_cov}
    % \vspace{-0.4cm}
\end{figure*}
\section{Illustration of Trajectory Covariance Prediction}
\label{ap:cov}

We present two intuitive examples to illustrate
\begin{enumerate*}[label=(\alph*)]
    \item how do different approaches model trajectory covariance,
    \item how does our model learn covariance in an affordable way.
\end{enumerate*}

\subsection{Modeling Trajectory Covariance}

We follow the definition of trajectory covariance used in the \acrshortpl{promp} \cite{paraschos2013probabilistic}, i.\,e., trajectory values of different \acrshortpl{dof} at different time steps are coupled and modeled in one multivariate distribution, which differs from the mixture models, such as GMM/GMR-DMP \cite{calinon2012statistical,yang2018robot}, where multiple Gaussian distribution components are used to cover the demonstrated trajectories' domain.
We illustrate an element-wise trajectory distribution example in Fig. \ref{fig:traj_full_cov}. 
The diagonal elements of the covariance matrix (colored in red) indicate the isotropic variance of the trajectory, while the remaining elements represent the covariance between time-steps and between different \acrshortpl{dof}. 
The original \acrshortpl{promp} and our \acrshortpl{pdmp} use a linear basis function model and can model the entire covariance matrix given $p(\bm{w}_g)\sim \mathcal{N}(\bm{w}_g|~\bm{\mu}_{w_g}, \bm{\Sigma}_{w_g})$. Recent approaches such as \acrshortpl{cnmp} \cite{seker2019conditional} and \acrshortpl{ndp} \cite{bahl2020neural} (RL part), however, can only model the isotropic trajectory variance (elements in red shadow in Fig. \ref{fig:traj_full_cov}). 
The correlation between different \acrshortpl{dof}, and between time steps is missing. Therefore, their predicted trajectory distribution cannot sample time and \acrshortpl{dof}-consistent new trajectories without further smoothness, which explains the reason for the noisy samples in Fig. \ref{fig:digit_cov}. 

\subsection{Make Full Covariance Learning Affordable}
\label{app:cov_pairs}
As briefly introduced in Section \ref{sbsec:nmp_model}, learning all time steps' covariance whose size is $TD \times TD$, requires a high resource demand. 
% The time complexities w.r.t the matrix's dimension of different numerical methods are between $O(n^{2.37})$ and $O(n^3)$ \cite{bientinesi2008families, van2011notes}. 
Such a high resource demand is often not feasible for a deep NN model, where multiple trajectories are often learned in a batched manner, and the computational graph shall be stored for back-propagation usage.
Therefore, to make the learning feasible, we randomly select a group of paired time steps, each pair having a $2D \times 2D$ covariance, and optimize the trajectory values' joint distribution. As illustrated in Fig. \ref{fig:cov_tp}, learning a distribution of paired time steps is equivalent to learning a sub-vector and a sub-matrix of the original ones. As such time pairs are randomly selected in different mini-batches, the original covariance matrix will be learned eventually.
\begin{figure*}[htbp]
    \centering
\vspace{-0.3cm}
\tikzset{ 
    table/.style={
        matrix of math nodes,
        row sep=-\pgflinewidth,
        column sep=-\pgflinewidth,
        nodes={rectangle,text width=3em,align=center},
        text depth=1.25ex,
        text height=2.5ex,
        nodes in empty cells,
        left delimiter=[,
        right delimiter={]},
        ampersand replacement=\&
    }
}
\usetikzlibrary{arrows,matrix,positioning}
    \begin{tikzpicture}
        % Origin mean
        \matrix [matrix of math nodes, left delimiter=[, right delimiter={]},text depth=1.25ex,
        text height=1.25ex,] (mean)
        {
            \mu_{1}\\ 
            \mu_{2}\\ 
            \mu_{3}\\ 
            \mu_{4}\\ 
            \mu_{5}\\ 
            \mu_{6}\\ 
        };
        \draw [dotted] (mean.west) -- (mean.east);
        \draw[rounded corners,ultra thick, draw=black, fill=blue, opacity=0.1] (mean-2-1.south west) rectangle (mean-1-1.north east);             
        \draw[rounded corners,ultra thick, draw=black, fill=blue, opacity=0.1] (mean-5-1.south west) rectangle (mean-4-1.north east);
        % Origin cov
        \matrix [matrix of math nodes, left delimiter=[, right delimiter={]}, text depth=1.25ex,
        text height=1.25ex,] (m) [right=0.5cm of mean]
        {
            \sigma_{1,1} & \sigma_{1,2} & \sigma_{1,3}\phantom{..} & \sigma_{1,4} & \sigma_{1,5} & \sigma_{1,6} \\ 
            \sigma_{2,1} & \sigma_{2,2} & \sigma_{2,3}\phantom{..} & \sigma_{2,4} & \sigma_{2,5} & \sigma_{2,6} \\ 
            \sigma_{3,1} & \sigma_{3,2} & \sigma_{3,3}\phantom{..} & \sigma_{3,4} & \sigma_{3,5} & \sigma_{3,6} \\ 
            \sigma_{4,1} & \sigma_{4,2} & \sigma_{4,3}\phantom{..} & \sigma_{4,4} & \sigma_{4,5} & \sigma_{4,6} \\ 
            \sigma_{5,1} & \sigma_{5,2} & \sigma_{5,3}\phantom{..} & \sigma_{5,4} & \sigma_{5,5} & \sigma_{5,6} \\ 
            \sigma_{6,1} & \sigma_{6,2} & \sigma_{6,3}\phantom{..} & \sigma_{6,4} & \sigma_{6,5} & \sigma_{6,6} \\ 
        };
        
        %fancy blue rectangle
        \draw[rounded corners,ultra thick, draw=black, fill=blue, opacity=0.1] (m-2-1.south west) rectangle (m-1-2.north east);             
        \draw[rounded corners,ultra thick, draw=black, fill=blue, opacity=0.1] (m-2-4.south west) rectangle (m-1-5.north east);
        \draw[rounded corners,ultra thick, draw=black, fill=blue, opacity=0.1] (m-5-1.south west) rectangle (m-4-2.north east);       
        \draw[rounded corners,ultra thick, draw=black, fill=blue, opacity=0.1] (m-5-4.south west) rectangle (m-4-5.north east);       
        \draw [dotted] (m.west) -- (m.east);
        \draw [dotted] (m.north) -- (m.south);
        
        % Small mean
        \matrix [matrix of math nodes, left delimiter=[, right delimiter={]},text depth=1.25ex,
        text height=1.25ex,] (mean_small) [right=2cm of m.east]
        {
            \mu_{1}\\ 
            \mu_{2}\\ 
            \mu_{4}\\ 
            \mu_{5}\\ 
        };
        % Small cov
        \matrix [matrix of math nodes, left delimiter=[, right delimiter={]},text depth=1.25ex,
        text height=1.25ex,] (m_small) [right=0.5cm of mean_small.east ]
        {
            \sigma_{1,1} & \sigma_{1,2} & \sigma_{1,4} & \sigma_{1,5} \\ 
            \sigma_{2,1} & \sigma_{2,2} & \sigma_{2,4} & \sigma_{2,5} \\ 
            \sigma_{4,1} & \sigma_{4,2} & \sigma_{4,4} & \sigma_{4,5} \\ 
            \sigma_{5,1} & \sigma_{5,2} & \sigma_{5,4} & \sigma_{5,5} \\ 
        };
        
        % Arrow
        \draw[-{Stealth[scale=1]}, ultra thick, color=\arrowblue] (m.east)+(1.5em, 0.0em) -- (22em, 0em);
    \end{tikzpicture}    
    \caption{An illustration of learning full covariance through paired time steps. \textbf{Left}: This example trajectory has coupled 3 time steps and 2 \acrshortpl{dof}. Here we omit the time-step and \acrshort{dof} labels for simplicity. \textbf{Right}: We randomly select 2 time steps, e.g. $t_1, t_2$, form up a pair, and learn their joint distribution, which is equivalent to learn a sub-vector and a sub-matrix of the original ones. 
    % Since such random selection is executed in every mini-batch during training, we eventually learn the original covariance using a relatively low cost.
    }
    \label{fig:cov_tp}
\end{figure*}
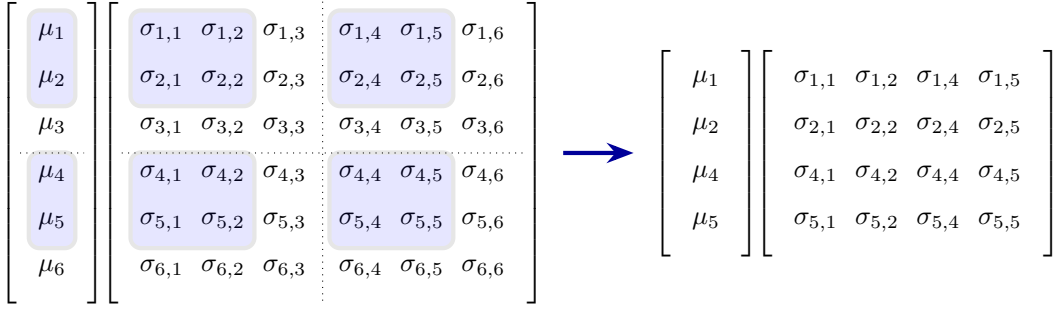

\section{Experiment Hyper-parameters Settings}
The hyper-parameters of our experiments and neural network architectures are listed in TABLE \ref{tab:one_digit_setting} - \ref{tab:robot_picking_setting} respectively.

%%%%%%%%%%%%%%%%%%%%%%%%%%%%%%%%%%%%%%%%%%%%%%%
% One digit
%%%%%%%%%%%%%%%%%%%%%%%%%%%%%%%%%%%%%%%%%%%%%%%
% \paragraph{Predict Digit Writing Trajectory Given One Image}
% We present the settings of different approaches in Table \ref{tab:one_digit_setting} and offer a supplementary result of our model with full covariance learning using all types of digits. 

\renewcommand{\arraystretch}{1.3}
\begin{table*}[ht]
    \footnotesize
    % \footnotesize
    \centering
    \caption{Experiment settings of writing digit given one original image.}
    \hspace{-0.2cm}
    \begin{tabular}{c|c|c|c}
        \toprule
        \textbf{Settings} & \centering \textbf{\acrshortpl{cnmp}}
        & \centering \textbf{NN-based \acrshortpl{dmp}} & \textbf{\acrshortpl{pdmp}}\\
        \hline
        Input obs. & \multicolumn{3}{c}{One original digit image, size: 40 $\times$ 40} \\
        \hline
        Predict & Traj. mean and std  & \acrshortpl{dmp} weights $+$ goal & \multicolumn{1}{c}{Mean $+$ cov of \acrshortpl{pdmp} weights} \\\hline
        \begin{tabular}{c}
            Form up \\
            trajectory 
        \end{tabular}& - & Numerical integration & \multicolumn{1}{c}{Use linear basis function in Eq.~(\ref{eq:idmp_pos})} \\
        \hline
        Loss func. & Traj. NLL &Traj. MSE & \multicolumn{1}{c}{Traj. LL-loss in Eq. (\ref{eq:rec_ll})} \\
        \hline
        Traj. sampling& From traj. distribution & $\times$ & \multicolumn{1}{c}{
        \begin{tabular}{@{}c@{}}First sample weights \\then use linear basis function Eq.~(\ref{eq:idmp_pos}) \end{tabular}}\\
        \hline
        No. of Basis &-&25 per \acrshort{dof}&\multicolumn{1}{c}{25 +1 (weights + goal) per \acrshort{dof}} \\
        \hline
        No. of \acrshortpl{dof} &\multicolumn{3}{c}{2 in task space} \\
        \hline
        Type of Basis &-&RBF&\multicolumn{1}{c}{RBF} \\
        \hline
        Hyper-params. &\multicolumn{3}{c}{\begin{tabular}{c}
            Optimizer: [Adam, learning rate: 2e-4, weights decay: 5e-5] \\
            Dim of latent obs.: 128. \acrshortpl{dmp}: [$\alpha_x$: 2, $\alpha$: 25, $\tau$: 3]              
        \end{tabular} }  \\
        \hline
        \begin{tabular}{cc|cc} Obs. encoder \\ Unc. encoder \\ in Fig. \ref{fig:nmp_structure}\end{tabular}&\multicolumn{3}{c}{
            \begin{tabular}{cc|cc}
                \multicolumn{4}{c}{\{~~CNN: [kernel: 5, stride: 1,  channel in: 1, channel out: 10] }   \\
                \multicolumn{4}{c}{Max pooling: [kernel: 2, stride: 2,  channel in: 10, channel out: 10] } \\
                \multicolumn{4}{c}{CNN: [kernel: 5, stride: 1,  channel in: 10, channel: out: 20, dropout: 0.5] } \\
                \multicolumn{4}{c}{Max pooling: [kernel: 2, stride: 2,  channel in: 20, channel out: 20]} \\
                \multicolumn{4}{c}{MLP: [input neurons: 980, hidden: [128, 128, 128],  output: 128]} ~\} \\
            \end{tabular}} \\
        \hline
        \begin{tabular}{cc|cc} Mean. dec. \\ Unc. dec. \\ in Fig. \ref{fig:nmp_structure}\end{tabular} &\multicolumn{3}{c}{
            \begin{tabular}{cc|cc}
                \multicolumn{4}{c}{\acrshortpl{pdmp} Mean: MLP[input: 128, hidden: [128, 128, 128],  output: 54]}\\
                % \multicolumn{4}{c}{\acrshortpl{pdmp} Std: MLP[input: 128, hidden: [128, 128, 128],  output: 54]}\\
                \multicolumn{4}{c}{\acrshortpl{pdmp} Cov: MLP[input: 128, hidden: [256, 256, 256, 256],  output: 1485]}\\
                \hline
                \multicolumn{4}{c}{\acrshortpl{cnmp} Mean: MLP[input: 128 + 1, hidden: [128, 128, 128],  output: 2]}\\
                % \multicolumn{4}{c}{\acrshortpl{cnmp} Std: MLP[input: 128 + 1, hidden: [128, 128, 128],  output: 2]}\\
                \multicolumn{4}{c}{NN-\acrshortpl{dmp} Mean: MLP[input: 128, hidden: [128, 128, 128],  output: 54]}\\
            \end{tabular}} \\
        \bottomrule
    \end{tabular}
    \label{tab:one_digit_setting}
    \vspace{-0.5cm}
\end{table*}

% \input{figure/tex/all_one_digit}

%%%%%%%%%%%%%%%%%%%%%%%%%%%%%%%%%%%%%%%%%%%%%%%
% Noisy digit
%%%%%%%%%%%%%%%%%%%%%%%%%%%%%%%%%%%%%%%%%%%%%%%
% \newpage
% \paragraph{Predict Digit Writing Trajectory Given Three Noisy Images}
% We present the setting of our experiment in Table \ref{tab:noisy_digit_setting}, and more sampled trajectories result in Fig. \ref{fig:noisy_digit_supplementary}.

\renewcommand{\arraystretch}{1.3}
\begin{table*}[ht]
    % \scriptsize
    \footnotesize
    \centering
    \caption{Experiment setting of writing digit given at most 3 noisy images.}
    \begin{tabular}{c|c}
        \toprule
        \textbf{Settings} & \textbf{\acrshortpl{pdmp}}\\
        \hline
        Input observation & One, two, or three noisy digit images generated from the original image \\
        \hline
        Predict & Mean and covariance of \acrshortpl{pdmp} weights \\
        \hline
        Form up trajectory 
        & Use linear basis function in Eq.~(\ref{eq:idmp_pos}) \\
        \hline
        Loss func. Traj. & LL-loss in Eq.~(\ref{eq:rec_ll}) \\
        \hline
        Traj. sampling& \begin{tabular}{@{}c@{}}First sample weights, then use linear basis function Eq.~(\ref{eq:idmp_pos}) \end{tabular}\\
        \hline
        No. of Basis and \acrshortpl{dof} &25 + 1 per \acrshort{dof}, 2 \acrshortpl{dof} in task space, type: RBF \\
        \hline
        Hyper-params. &\begin{tabular}{c}
             Optimizer: [Adam, learning rate: 2e-4, weights decay: 5e-5] \\
             Gradient clipping norm: 20.\\
             Dim of latent obs.: 128. \acrshortpl{dmp}: [$\alpha_x$: 2, $\alpha$: 25, $\tau$: 3]
        \end{tabular} \\
        \hline
        Encoders and Decoders & Same as Table \ref{tab:one_digit_setting}\\
        \bottomrule
    \end{tabular}
    \label{tab:noisy_digit_setting}
    \vspace{-0.5cm}
\end{table*}

% \input{figure/tex/noisy_digit_additional}
%%%%%%%%%%%%%%%%%%%%%%%%%%%%%%%%%%%%%%%%%%%%%%%
% Digit replan
%%%%%%%%%%%%%%%%%%%%%%%%%%%%%%%%%%%%%%%%%%%%%%%
% \clearpage
% \newpage
% \paragraph{Writing a Digit in Three Steps}
% The setting of our experiment can be found in Table \ref{tab:digit_relanning_setting}. A supplementary result to Fig. \ref{fig:digit_replan} can be found in Fig. \ref{fig:digit_replan_supplementary}, where we show examples of all 10 digits' 3-step writing and replanning procedure.

\renewcommand{\arraystretch}{1.3}
\begin{table*}[ht]
    % \scriptsize
    \footnotesize
    \centering
    \caption{Experiment setting of writing digit in 3 steps.}
    \begin{tabular}{c|c}
        \toprule
        \textbf{Settings} & \textbf{\acrshortpl{pdmp} using sub-datasets containing 1 type of digit}\\
        \hline
        Input observation & One, two, or three noisy digit images generated from the original image \\
        \hline
        Predict & Mean and covariance of \acrshortpl{pdmp} weights \\\hline
        \begin{tabular}{c}
            Form up \\
            trajectory 
        \end{tabular}& \begin{tabular}{c}
            Use linear basis function in Eq.~(\ref{eq:idmp_pos})\\
            Use the execution of the previous prediction \\
            at 0\%, 25\% and 50\% respectively as boundary condition 
        \end{tabular}  \\
        \hline
        Loss func. Traj. & LL-loss in Eq.~(\ref{eq:rec_ll}) \\
        \hline
        No. of Basis &25 + 1 per \acrshort{dof} \\
        \hline
        No. of \acrshortpl{dof} &\multicolumn{1}{c}{2 in task space} \\
        \hline
        Type of Basis &RBF \\
        \hline
        Hyper-params. &\begin{tabular}{c}
             Optimizer: [Adam, learning rate: 2e-4, weights decay: 5e-5] \\
             Gradient clipping norm: 20.\\
             Dim of latent obs.: 128. \acrshortpl{dmp}: [$\alpha_x$: 2, $\alpha$: 25, $\tau$: 3]
        \end{tabular} \\
        \hline
        Encoders and Decoders & Same as Table \ref{tab:one_digit_setting}\\
        \bottomrule
    \end{tabular}
    \label{tab:digit_relanning_setting}
    \vspace{-0.5cm}
\end{table*}

% \input{figure/tex/digit_replan_supplementary}

%%%%%%%%%%%%%%%%%%%%%%%%%%%%%%%%%%%%%%%%%%%%%%%
% Robot Pushing
%%%%%%%%%%%%%%%%%%%%%%%%%%%%%%%%%%%%%%%%%%%%%%%
% \clearpage
% \newpage
% \paragraph{Robot Pushing}
% \input{figure/tex/robot_push_illustration}
% We illustrate the robot pushing task in Fig. \ref{fig:robot_push_illustration}, and the settings of different approaches in Table \ref{tab:robot_push_setting}. We show a comparison of the pushing trajectory between our model and \acrshort{cnmp} in Fig. \ref{fig:cnmp_push}. It is worth mentioning that the robot pushing trajectory in our settings is a concatenation of multiple trajectory segments. Each segment takes either the mean or sample of the predicted trajectory distribution at each replanning step. The two exceptions are \acrshort{pdmp} (entire) and Behaviour Cloning Net. The former predicts the entire trajectory without replanning while the latter predicts a trajectory value at each single time step given the current robot and box state. 
\vspace{0.3cm}
\renewcommand{\arraystretch}{1.4}
\begin{table*}[ht]
    % \scriptsize
    \footnotesize
    \centering
    \caption{Experiment settings of robot pushing task.}
    % \centering
    % \begin{subtable}[h]{\textwidth}
        % \hspace{1cm}
        \begin{tabular}{c|c|c}
            \toprule
            \textbf{Settings} & \centering \textbf{\acrshortpl{cnmp}}
            & \textbf{\acrshortpl{pdmp}}\\
            \hline
            Input obs. & \multicolumn{2}{c}{The box and robot's actual state in the past 0.1 second} \\
            \hline
            Predict & Traj. mean and std of next 0.5s & \multicolumn{1}{c}{Mean $+$ cov of \acrshortpl{pdmp} weights in the next 0.5s} \\\hline
            \begin{tabular}{c}
            Form up \\
            trajectory 
        \end{tabular}& - & \multicolumn{1}{c}{Use linear basis function in Eq.~(\ref{eq:idmp_pos})} \\
            \hline
            Loss func. & Traj. NLL & \multicolumn{1}{c}{Traj. LL-loss in Eq.~(\ref{eq:rec_ll})} \\
            \hline
            Traj. sampling& From traj. distribution &
            \begin{tabular}{@{}c@{}}First sample weights \\then use linear basis function Eq.~(\ref{eq:idmp_pos}) \end{tabular}\\
            \hline
            No. of Basis &-& \multicolumn{1}{c}{25 + 1 per \acrshort{dof}} \\
            \hline
            No. of \acrshortpl{dof} &\multicolumn{2}{c}{2 in task space} \\
            \hline
            Type of Basis &-&\multicolumn{1}{c}{RBF} \\
            \hline
            Hyper-params. &\multicolumn{2}{c}{\begin{tabular}{c}
                Optimizer: [Adam, learning rate: 2e-4, weights decay: 5e-5] \\
                Dim of latent obs.: 128. \acrshortpl{dmp}: [$\alpha_x$: 0.5, $\alpha$: 25, $\tau$: 0.5]              
            \end{tabular} }  \\            
            \bottomrule
        \end{tabular}
    % \end{subtable}
    \vspace{0.5cm}
    % \begin{subtable}[h]{\textwidth}
        % \hspace{1cm}
        \begin{tabular}{c|c|c}
            \toprule
            \textbf{Settings} & \textbf{\acrshortpl{pdmp} no replanning} & \textbf{Behaviour Cloning}\\
            \hline
            Input obs. & The initial box and robot state & The latest actual box and robot state~~\\
            \hline
            Predict & Mean $+$ cov of \acrshortpl{pdmp} weights of the entire traj. & Robot's desired one-step trajectory \\\hline
            \begin{tabular}{c}
            Form up \\
            trajectory 
        \end{tabular} & Use linear basis function in Eq.~(\ref{eq:idmp_pos}) & - \\
            \hline
            Loss func. & Traj. LL-loss in Eq. (\ref{eq:rec_ll}) & MSE loss\\
            \hline
            No. of Basis & 25 + 1 per \acrshort{dof}&-\\
            \hline
            No. of \acrshortpl{dof} &\multicolumn{2}{c}{2 in task space} \\
            \hline
            Type of Basis & RBF & - \\
            \hline
            Hyper-params. &\multicolumn{2}{c}{\begin{tabular}{c}
                Optimizer: [Adam, learning rate: 2e-4, weights decay: 5e-5] \\
                \acrshortpl{dmp}: [$\alpha_x$: 3, $\alpha$: 25, $\tau$: 3]              
            \end{tabular} }  \\       
            \hline
            \begin{tabular}{c}
                Encoders,  \\
                decoders, \\
                and \\
                other NNs
            \end{tabular} &\multicolumn{2}{c}{
                \begin{tabular}{cc|cc}
                    \multicolumn{4}{c}{\acrshortpl{pdmp} Obs./Unc. enc.: MLP[input neuron: 8, hidden:[128, 128, 128], output: 128]}   \\
                    \multicolumn{4}{c}{\acrshortpl{cnmp} Obs. enc.: MLP[input neuron: 8, hidden:[128, 128, 128], output: 128]}   \\
                    \hline
                    \multicolumn{4}{c}{\acrshortpl{pdmp} Mean. dec.: MLP[input neuron: 128, hidden:[128, 128, 128], output: 52]}   \\
                    \multicolumn{4}{c}{\acrshortpl{pdmp} Cov. dec.: MLP[input neuron: 128, hidden:[128, 128, 128], output: 1326]}   \\
                    \multicolumn{4}{c}{\acrshortpl{cnmp} Mean/Std dec.: MLP[input neuron: 128+1, hidden:[128, 128, 128], output: 2]}   \\
                    \hline
                    \multicolumn{4}{c}{Behaviour Cloning Net: MLP[input neuron: 6$^*$, hidden:[128, 128, 128], output: 2]}   \\
                    \multicolumn{4}{c}{$^*$We found that not offering the robot actual velocity (2-dim) increases the performance.}   \\
                \end{tabular}} \\
            \bottomrule
        \end{tabular}
    % \end{subtable}
    \label{tab:robot_push_setting}
    % \vspace{-0.5cm}
\end{table*}

% \input{figure/tex/cnmp_jumps}

%%%%%%%%%%%%%%%%%%%%%%%%%%%%%%%%%%%%%%%%%%%%%%%
% Robot Picking
%%%%%%%%%%%%%%%%%%%%%%%%%%%%%%%%%%%%%%%%%%%%%%%
% \clearpage
% \newpage
% \subsection{Robot Picking}

% The setting of our experiment can be found in Table \ref{tab:robot_picking_setting}, and we offer supplementary plots of the joint trajectory of each \acrshort{dof} in Fig. \ref{fig:robot_pick_additional}.
\renewcommand{\arraystretch}{1.5}
\begin{table*}[ht]
    % \scriptsize
    \footnotesize
    \centering
    \caption{Experiment settings of robot picking task.}
    \begin{tabular}{c|c}
        \toprule
        \textbf{Settings} & \textbf{\acrshortpl{pdmp}}\\
        \hline
        Input observation & The object and robot state at the 0, 1, 2, 3 seconds  \\
        \hline
        Predict & Mean and covariance of \acrshortpl{pdmp} weights \\\hline
        Post-process& Use linear basis function in Eq.~(\ref{eq:idmp_pos}) \\
        \hline
        Loss func. Traj. & LL-loss in Eq.~(\ref{eq:rec_ll}) \\
        \hline
        Traj. sampling& \begin{tabular}{@{}c@{}}First sample weights \\then use linear basis function Eq.~(\ref{eq:idmp_pos}) \end{tabular}\\
        \hline
        No. of Basis & 9 + 1 per \acrshort{dof} \\
        \hline
        No. of \acrshortpl{dof} & 7 in joint space\\
        \hline
        Type of Basis &RBF \\
        \hline
        Hyper-params. &\begin{tabular}{c}
            Optimizer: [Adam, learning rate: 2e-4, weights decay: 5e-5] \\
            Dim of latent obs.: 128. \acrshortpl{dmp}: [$\alpha_x$: 3, $\alpha$: 25, $\tau$: 4]              
            \end{tabular}  \\         
        \hline
        Encoders & Obs./Unc. encoder: MLP[input neurons: 4, hidden: [64, 64, 64],  output: 64] \\
        \hline
        Decoders &\begin{tabular}{c}
                        \acrshortpl{pdmp} Mean: MLP[input: 64, hidden: [64, 64, 64],  output: 70]\\
                        \acrshortpl{pdmp} Cov: MLP[input: 64, hidden: [64, 64, 64],  output: 2415]\\
            \end{tabular} \\
        \bottomrule
    \end{tabular}
    \label{tab:robot_picking_setting}
\end{table*}

}

% \bibliographystyle{IEEEtran}
% \bibliography{reference}

\end{document}